\documentclass[letterpaper, 10pt, conference]{ieeeconf} 
\usepackage[english]{babel}
\usepackage{subcaption}

\usepackage{amsmath}
\usepackage{graphicx}
\usepackage[colorlinks=true, allcolors=blue]{hyperref}

\title{Evaluating the precision of the HTC VIVE\textregistered\ Ultimate Tracker with robotic and human movements under varied environmental conditions}

\author{Julian Kulozik, Nathanaël Jarrassé\\
\small{Sorbonne Université, CNRS, INSERM, Institute for Intelligent Systems and Robotics (ISIR), Paris, France.}\\
Email: \textit{kulozik@isir.upmc.fr}}%

\begin{document}
\maketitle

\begin{abstract}
The HTC VIVE\textregistered\ Ultimate Tracker, utilizing inside-out tracking with internal stereo cameras providing 6 DoF tracking without external cameras, offers a cost-efficient and straightforward setup for motion tracking. Initially designed for the gaming and VR industry, we explored its application beyond VR, providing source code for data capturing in both C++ and Python without requiring a VR headset \cite{MyRepo}. This study is the first to evaluate the tracker's precision across various experimental scenarios, including tracking robotic movements, natural human motion, and dynamic sports activities.

To assess the robustness of the tracking precision, we employed a robotic arm as a precise and repeatable source of motion.%, allowing us to independently and comprehensively analyze the effects of various perturbations. 
Using the OptiTrack system as a reference, we conducted tests under varying experimental conditions: lighting, movement velocity, environmental changes caused by displacing objects in the scene, and human movement in front of the trackers, as well as varying the displacement size relative to the calibration center.

On average, the HTC VIVE Ultimate Tracker achieved a precision of 4.98 mm $\pm$ 4 mm across various conditions. %In best case scenario %(constant good lighting, small range of motion) the precision was measured to be 2.59 mm $\pm$ 0.81 mm 
The most critical factors affecting accuracy were lighting conditions, movement velocity, and range of motion relative to the calibration center. %Low-light conditions and large displacements from the calibration center significantly degraded precision, while moderate environmental perturbations had a smaller but still noticeable impact.

For practical evaluation, we captured human movements with 5 trackers in realistic motion capture scenarios. Our findings indicate sufficient precision for capturing human movements, validated through two tasks: a low-dynamic pick-and-place task and high-dynamic fencing movements performed by an elite athlete. %In these tests, we placed the maximum supported five trackers at different regions of interest on the participant's body. While rotations are well-measurable by IMUs, the need for accurate translation measurement remains critical, and the VIVE Tracker contributes effectively in this aspect.

Even though its precision is lower than that of conventional fixed-camera-based motion capture systems and its performance is influenced by several factors, the HTC VIVE Ultimate Tracker demonstrates adequate accuracy for a variety of motion tracking applications. Its ability to capture human or object movements outside of VR or MOCAP environments makes it particularly versatile. 
%Additionally, the VIVE Tracker can be an excellent alternative when fixed cameras cannot be set up, space constraints limit camera field of view, or when recording needs to be conducted in multiple locations. This flexibility broadens its usability across a wide range of tracking scenarios.
\end{abstract}

\section{Introduction}

Motion capture systems play a critical role in various applications, ranging from virtual reality (VR) and gaming to bio mechanics, robotics, and human-computer interaction \cite{MocapReview}. These systems are designed to accurately track and record the movement of objects or individuals within a defined space. Traditional motion capture systems, camera based with reflective markers, such as Vicon\textregistered\ or OptiTrack\textregistered\, are renowned for their high precision and reliability. However, they require extensive setup, including multiple external cameras and complex calibration procedures, which can be both time-consuming and costly. Additionally, they demand rather large spaces to ensure an adequate field of view for the cameras, further complicating their deployment in smaller or more constrained environments.

In recent years, advancements in consumer-grade tracking technologies have opened new alternatives for more accessible and cost-effective motion capture solutions. Since November 2023, one such technology is the VIVE Ultimate Tracker, which employs inside-out tracking. Unlike traditional systems, the VIVE Ultimate Tracker does not rely on external cameras for tracking but instead uses embedded stereo cameras and SLAM algorithms to track itself within a self-created map. This independence from external infrastructure significantly reduces the overall cost and complexity of the motion capture setup, making it an attractive option for a wide range of applications.

While much of the current research is moving towards markerless motion capture solutions, particularly for biomechanical applications \cite{Mundermann2006,desmarais2021,Iseki2023}, these approaches still require external cameras, which can pose similar limitations in terms of space and setup complexity.

The primary objective of this study is to evaluate the measurement accuracy of the VIVE Ultimate Tracker, which was developed for virtual reality (VR) applications, with inside-out tracking in comparison to a field-standard optical motion capture system with multiple fixed cameras and reflective markers (here the OptiTrack system). By benchmarking the VIVE Tracker, we aim to determine its feasibility for movement tracking. This comparison is crucial as it will highlight the strengths and limitations of using consumer-grade VR trackers for real-world applications beyond their intended VR context.

Current research in motion capture predominantly focuses on the use of high-end systems like Vicon and OptiTrack, which are well-documented for their precision and robustness
\cite{Furtado2019,OptiTrackWebsite,ViconWebsite}. However, there is a growing interest in exploring more affordable and versatile alternatives that can deliver comparable performance. Previous studies have evaluated the HTC VIVE Tracker 3.0—a previous generation of VIVE trackers designed for VR that utilizes outside-in tracking technology, relying on external base stations for positioning—against the Vicon system, showing promising results in terms of accuracy and application potential \cite{Merker2023}. Our study extends this line of research by introducing the VIVE Ultimate Tracker, which uses inside-out tracking technology, and comparing it with the OptiTrack system, considering various environmental factors and movement forms and velocities.

In our experimental setup, a robot was programmed to perform various movements, including linear trajectories and circular patterns, at different speeds and different sizes of displacement. These movements were tracked simultaneously using the VIVE Ultimate Tracker and the OptiTrack system. We captured the data using custom Python code and analyzed the data to assess the accuracy of the VIVE Tracker. However, we focused our analysis on translation tracking rather than orientation tracking, as the latter is adequately handled by IMUs. Additionally, we investigated the impact of environmental conditions, such as lighting variations and changes in the tracking environment after creating the initial map, on the precision of the VIVE Tracker.

\section{Materials and Methods}

\subsection{Experimental Setup}

In our experimental setup, a robot was programmed to perform a variety of movements, including linear trajectories and circular patterns, at different speeds and displacement sizes. These motions were simultaneously tracked by both the VIVE Ultimate Tracker and the OptiTrack system. Positional data were captured using custom Python code\cite{MyRepo}, allowing us to evaluate the accuracy of the VIVE Tracker. Additionally, we examined how environmental factors, such as lighting variations and changes in the tracking environment post-mapping, influenced the precision of the VIVE Tracker.

For our experimental setup, we employed the KUKA LWR4+ 7 DoF robotic arm in combination with a 5-camera OptiTrack system, consisting of two PrimeX 13W cameras and three Prime13 cameras. The decision to use the OptiTrack system, rather than relying solely on the robot’s direct kinematics, was made to ensure that the data collected would be accurate and consistent, allowing for reliable comparisons between robotic and human movement experiments.

The OptiTrack cameras, including the PrimeX 13W and Prime13, offer a 3D accuracy ranging from ±0.20 mm to ±0.30 mm and were configured to capture data at a frame rate of 240 FPS. The PrimeX 13W cameras provide a field of view (FOV) of 82° horizontal and 70° vertical, with a tracking range of up to 9 meters for passive markers. In contrast, the Prime13 cameras have a tracking range of up to 16 meters, with a slightly narrower FOV of 56° horizontal and 46° vertical \cite{OptiTrackWebsite}. In our setup, the cameras were mounted on the ceiling, positioned between 3 meters and 5 meters away from the robot's end-effector.

\begin{figure}[!h]
\centering
\includegraphics[width=0.45\textwidth]{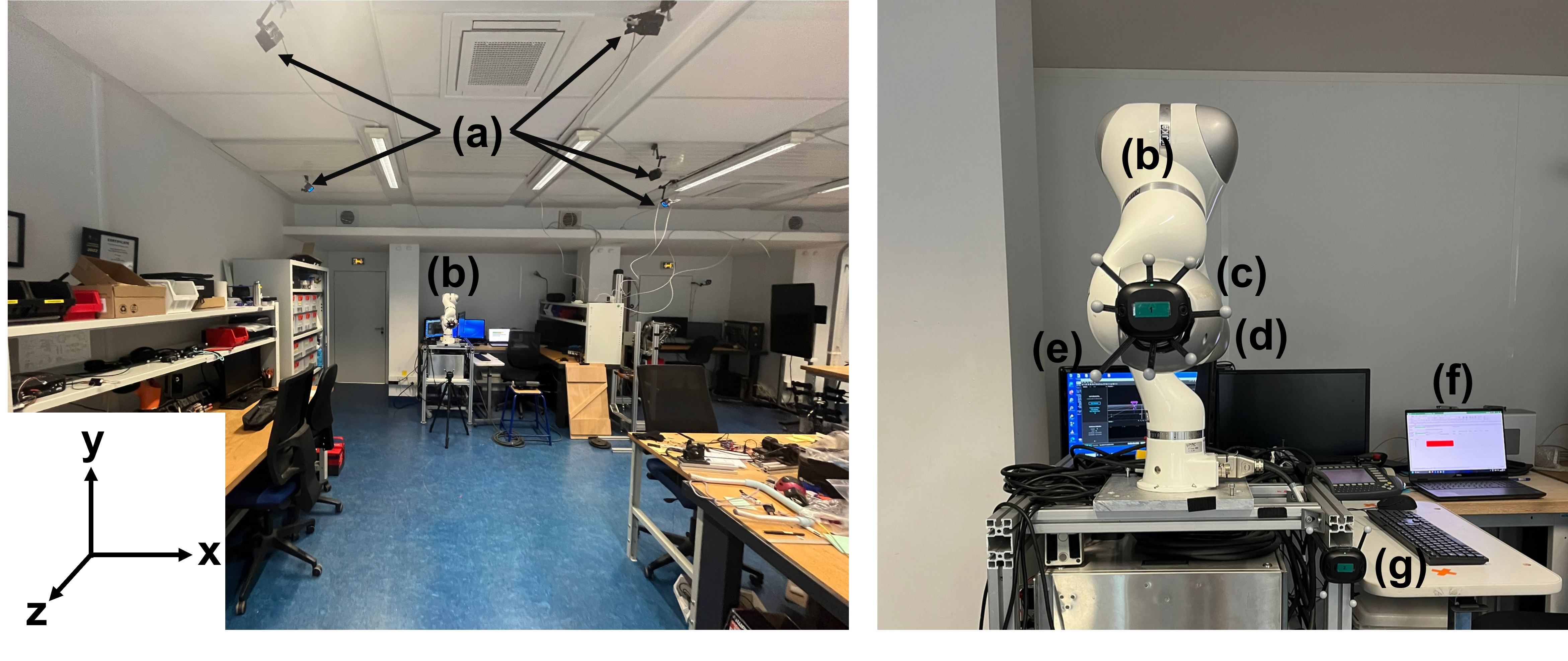}
\caption{\label{fig::ExpSetupKUKA} Experimental Setup KUKA robot. (a) OptiTrack cameras attached to the ceiling facing the robot, (b) KUKA LWR 4+ robot, (c) VIVE Ultimate Tracker, (d) OptiTrack marker set attached to the VIVE tracker, (e) Computer capturing OptiTrack data, (f) computer capturing VIVE data and receiving data from (e), (g) second OptiTrack-VIVE marker set, attached to the robot's movable base.}
\end{figure}

\subsection{Procedure} \label{Procedure}

\begin{figure}[!h]
\centering
\includegraphics[width=0.45\textwidth]{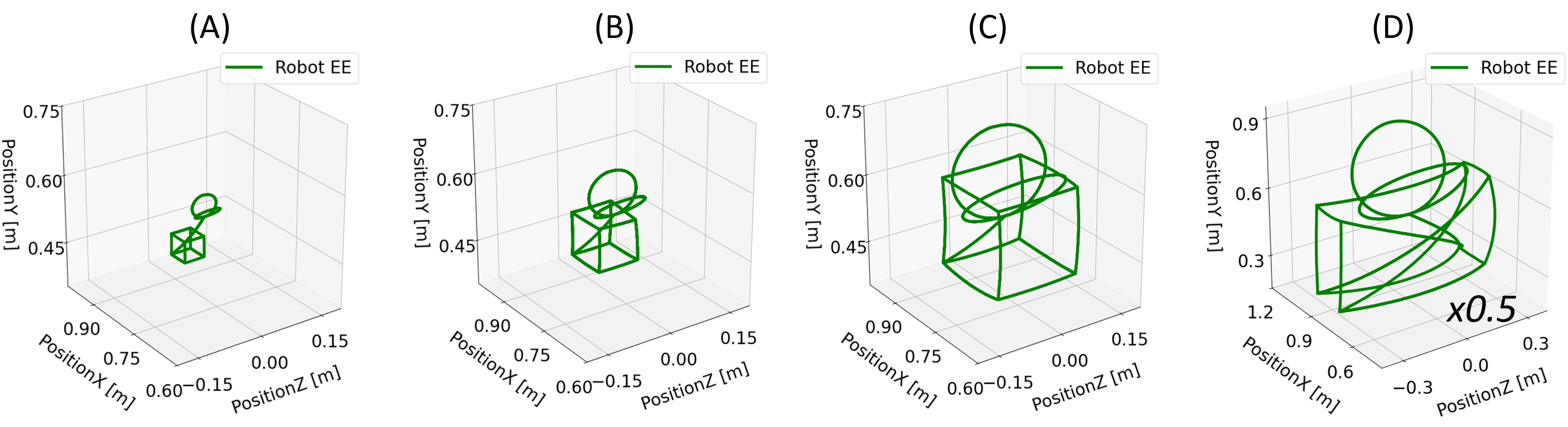}
\caption{\label{fig::RobotMovements} Four pre-programmed robotic movement patterns. (A) 5cm, (B) 10 cm, (C) 20 cm, (D) 80 x 40 x 20}
\end{figure}

To ensure a reproducible setup, we defined four different movement patterns for the robot to perform. These movement patterns are shown in Figure \ref{fig::RobotMovements}. On the very left, it's the smallest movement with a cube size and circle diameter of around 5 cm (A), 10 cm (B), and 20 cm (C). In (D), the edges of the movement are around 80 cm x 40 cm x 20 cm, which is about the robot's workspace.

Since the robot is just a source of reproducible movement and we don't need it to perform perfect cube movements nor perfect circles, we decided to joint control the robot to maximize its capability for high velocities. By controlling the redundant 7 Dof robot like this, we could ensure complete repeatability since no online kinematic calculations were done, and we could more easily exhibit the robot's dynamic capabilities.

The velocity limits for each displacement size varied due to different constraints. For very small movements, the limiting factor was the jerk caused by strong accelerations over short distances. For larger movements, the robot's own weight, positioned further from the center, was limited by its maximal torque.

For a displacement of 5 cm (Pattern A), the feasible peak velocity for the cubes was 20 cm/s, resulting in an average velocity of 10 cm/s and a constant maximum velocity for the circles of also 10 cm/s. This average was calculated considering the linear velocity profile, which ramps up from 0 to 20 cm/s and then back down to 0 for the linear movements. For a 10 cm displacement (Pattern B), the robot could achieve maximum velocities of up to 80 cm/s. For a 20 cm displacement (Pattern C) and the entire workspace movement (Pattern D), the maximum feasible velocity was 40 cm/s for the cube, so 20 cm/s average and a constant 20 cm/s for the circle.

The experimental setup involved varying lighting conditions using both the room's ceiling lights and an additional dimmable LED light source. The ceiling lights provided general illumination across the entire room, while the LED light source offered adjustable, localized lighting. These lights were positioned to allow for controlled changes in the room’s overall brightness. The LED light was capable of being dimmed from a very low brightness level, creating a dim environment, to full brightness, simulating more intense lighting conditions. This setup enabled us to simulate a range of lighting scenarios from well-lit to low-light environments to observe the impact on the tracking performance of the VIVE system.

The experimental setup was designed to evaluate the impact of environmental perturbations on the VIVE tracker’s accuracy. The setup included movable objects, such as chairs and various items, that were strategically positioned around the tracking area. These objects could be rearranged during the experiment to simulate changes in the environment. Additionally, the setup allowed for human activity within the tracking field, with a designated path for a person to walk through the scene in front of the tracker's built-in cameras. 

\subsection{Data and Statistical Analysis}

\paragraph{Calibration}
\label{CalibrationDiscussion}
In our initial step, we calculated the transformation between the VIVE frame and the OptiTrack frame. To do this, we first needed to identify the VIVE tracker's center of measurement. Without specific information from the HTC VIVE support and development team, we determined this center experimentally.

\begin{figure}
\centering
\includegraphics[width=0.45\textwidth]{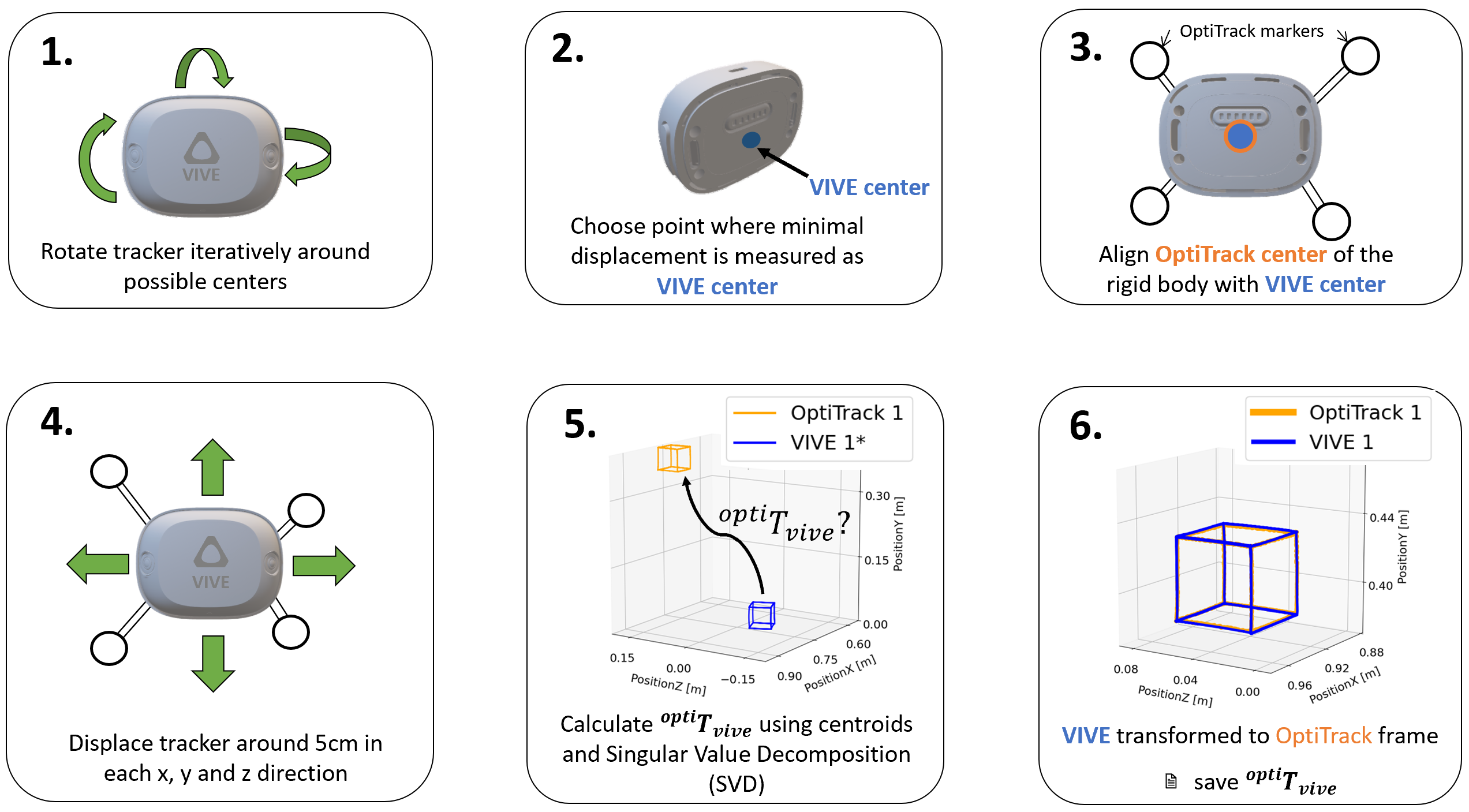}
\caption{\label{fig::CalibProcess} Calibration process. (1) Rotating tracker around different centers of rotation, (2) identifying center with smallest relative displacement measured, (3) adding OptiTrack markers and aligning OptiTrack center of created rigid body with identified VIVE center, (4) displace tracker in a cube pattern around 5 cm in each direction, (5) calculate transformation between VIVE and OptiTrack frame, (6) transform VIVE data and store transformation matrix.}
\end{figure}

First, we rotate the tracker iteratively around possible centers (Fig \ref{fig::CalibProcess}, Step 1). We do this through the teach panel of the KUKA robot where we can define the exact position of the tool center point manually. We then let the robot perform pure rotations around its tool center. By identifying the position of minimal displacement during these rotations around different tool centers, we define this position as the VIVE center (Fig \ref{fig::CalibProcess}, Step 2). This position is in fact around the center of the bottom of the tracker. By adding reflective markers to the outside of the tracker we can create a rigid body in the OptiTrack software. The center of this rigid body we then align manually within the OptiTrack software in all three dimensions with the identified center of the VIVE frame (Fig \ref{fig::CalibProcess}, Step 3). In a next step the robot displaces the tracker in a cube pattern with a cube size of 5 cm and at a velocity of 2.5 cm/s (Fig \ref{fig::CalibProcess}, Step 4). During this we capture both the OptiTrack's and the VIVE's positional data. Each being in their own frame, which is shown in Figure \ref{fig::CalibProcess} in step 5 where \textit{VIVE 1*} is the VIVE trackers position in the VIVE frame and \textit{OptiTrack 1} is the OptiTrack rigid body position in the OptiTrack frame. To calculate the transformation between the two frames we are using a commonly know method of centroids and Singular Value Decomposition (SVD) to both find the translation and the rotation between the two measurements \cite{Arun1987}. This method is further explained in Appendix \ref{Registration Method}. In Step 6 in Figure \ref{fig::CalibProcess}, we can see the well matched OptiTrack and VIVE positional data with a spatial distance of 0.80 mm $\pm$ 0.34 mm. The transformation matrix defined as $^{opti}T_{vive}$ is finally stored and used to transform the VIVE data to the OptiTrack frame. This calibration was re-done before each set of experiments. 

However, it is important to note that the calibration results were not always consistent. In some instances, significant scaling issues were observed, resulting in larger measurement errors. When this occurred, we re-scanned the room following the VIVE software instructions and repeated the calibration process until no visible scaling issues were detected.

\paragraph{Sampling rate}

To determine the unknown sampling rate of the VIVE Ultimate Tracker, we tested the behavior of the tracker at different sampling rates. We initially used our Python script to read out the sensor data, but we were limited to a maximum sampling rate of around 220 Hz. Within this range, we did not observe any typical sampling limits or repeated values.

To further investigate, we employed our C++ code (available at \cite{MyRepo}), allowing us to sample at significantly higher rates, up to 5000 Hz. At this higher sampling rate, we still did not obtain identical consecutive values. Instead, we observed that the tracker employs trajectory prediction to generate the next value at a visibly non-constant rate, predicting the direction of movement.

\begin{figure}[!h]
\centering
\includegraphics[width=0.47\textwidth]{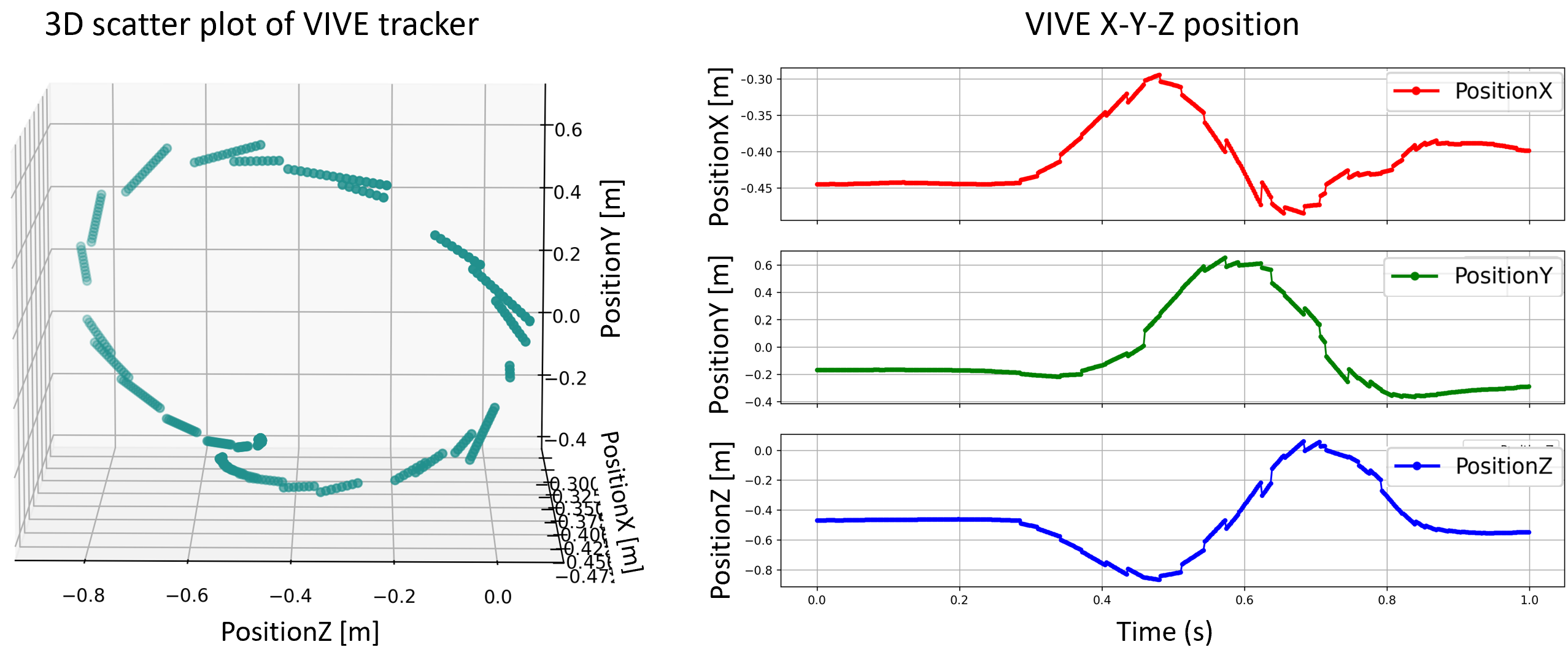}
\caption{\label{fig:500HZCircle}
Left: 3D scatter plot showing the trajectory of the VIVE tracker during a high-speed circular movement, demonstrating the effect of trajectory prediction at a sampling rate of 500 Hz. Right: Position of the VIVE tracker over time in the X, Y, and Z axes, highlighting the non-constant rate of value prediction and the trajectory generation.}
\end{figure}

This phenomenon is clearly illustrated in Figure \ref{fig:500HZCircle}. For this experiment, the tracker was held in the hand and a rapid circular movement was performed to maximize the observable effects of this trajectory prediction mechanism.

After gradually increasing the sampling rate from 30 Hz to 5000 Hz, we determined that the true sampling rate, without any trajectory predictions, is approximately 120 Hz. Based on this finding, we standardized our experiments to use this 120 Hz sampling rate moving forward.

\paragraph{Latency}
To estimate the latency of the VIVE Ultimate Tracker, we considered the latency's of the components in our experimental setup. The OptiTrack cameras used have a known latency of 4.2 ms. Additionally, the ping between the two computers (one capturing the OptiTrack data and the other capturing the VIVE data and receiving the OptiTrack data via OSC) was measured to be approximately 5 ms.

By analyzing the synchronized position data from both systems, we observed that the position jumps appeared without any noticeable shift. This synchronization suggests that the latency of the VIVE Ultimate Tracker is approximately 10 ms. This estimate accounts for the cumulative delays from both the OptiTrack system and the network communication between the computers.

\paragraph{Statistics}
For the statistical analysis, we primarily used ANOVA, given the large sample sizes (n $>$ 2000) in our datasets. Even for data that did not strictly adhere to normality, ANOVA was deemed appropriate due to the robustness of the method with large datasets. We performed either one-way or two-way ANOVA to assess the effects of different variables on tracking error. When significant effects were identified, post-hoc tests were conducted to determine specific group differences.

All statistical analyses were conducted using JASP (version 0.18.3.0).

\section{Results}

\subsection{Impact of Lighting Conditions}

To investigate the influence of lighting on tracker precision, we designed an experiment using a robotic arm performing repetitive movements following Pattern B (Figure \ref{fig::3Dplot_DisplaceRobot_grid}), which involved a 10 cm displacement at an average velocity of 10 cm/s. The lighting conditions were manipulated using the room's ceiling lights and an additional dimmable LED light source. During each trial, we exposed the tracker to varying lighting scenarios to assess the impact on tracking accuracy. Initially, the tracker was tested under consistent bright lighting, where the ceiling lights were kept on at full brightness without any additional lighting. We then tested under low lighting by turning off the ceiling lights and dimming the LED light source to create a dim environment. Dynamic lighting changes were introduced by keeping the ceiling lights on while varying the LED brightness between low and high levels. Additionally, a dark-to-bright transition was conducted, starting with the LED at very low brightness and gradually increasing to full brightness during the movement.

\begin{table}[h]
\centering
{
\begin{tabular}{lrrrr}
& Mean & Std. Deviation & Max \\
\hline
Constant Bright & $2.574$ & $1.067$ & $5.6$ \\
Constant Low & $10.635$ & $3.213$& $19.2$ \\
Bright to Brighter & $3.734$ & $1.615$ & $12.1$ \\
Low to Bright & $6.422$ & $2.386$ & $13.5$ \\
\end{tabular}
\caption{\label{tab:Table_differentLight} Measured spatial distances between VIVE and OptiTrack [mm] under varying lighting conditions.}
}
\end{table}

\begin{figure}[!h]
\centering
\includegraphics[width=0.3\textwidth]{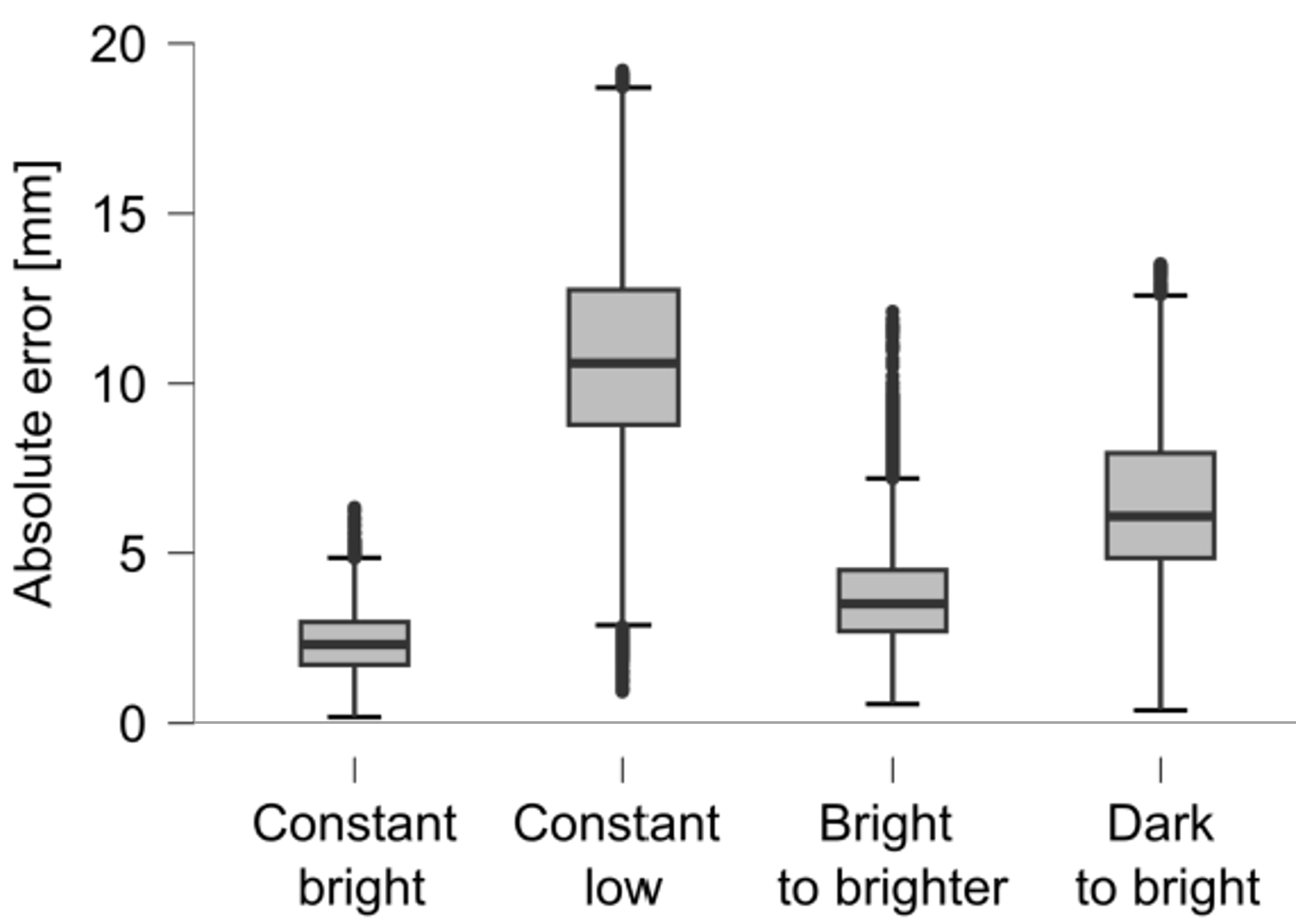}
\caption{\label{fig::Boxplot_differentLight} Boxplot of the absolute error [mm] at different lighting conditions at a 10 cm range of motion.}
\end{figure}

The results in \ref{tab:Table_differentLight} and \ref{fig::Boxplot_differentLight} indicate that low light conditions significantly degrade the tracker's precision. Continuous changes in lighting however barely negatively impact precision. Natural light was excluded from this study due to its poor repeatability, although its significant impact was noted. Additionally, using the tracker outdoors was not feasible, likely because the tracker relies on environmental references that need to be relatively close.

\subsection{Impact of Environmental Perturbations}

To evaluate the effect of environmental perturbations on the tracker’s performance, we introduced changes during data capture, such as rearranging furniture and having a person walk through the tracker’s field of view. These perturbations were designed to assess how dynamic environmental changes influenced the tracker’s accuracy.

\begin{figure}[!h]
\centering
\includegraphics[width=0.3\textwidth]{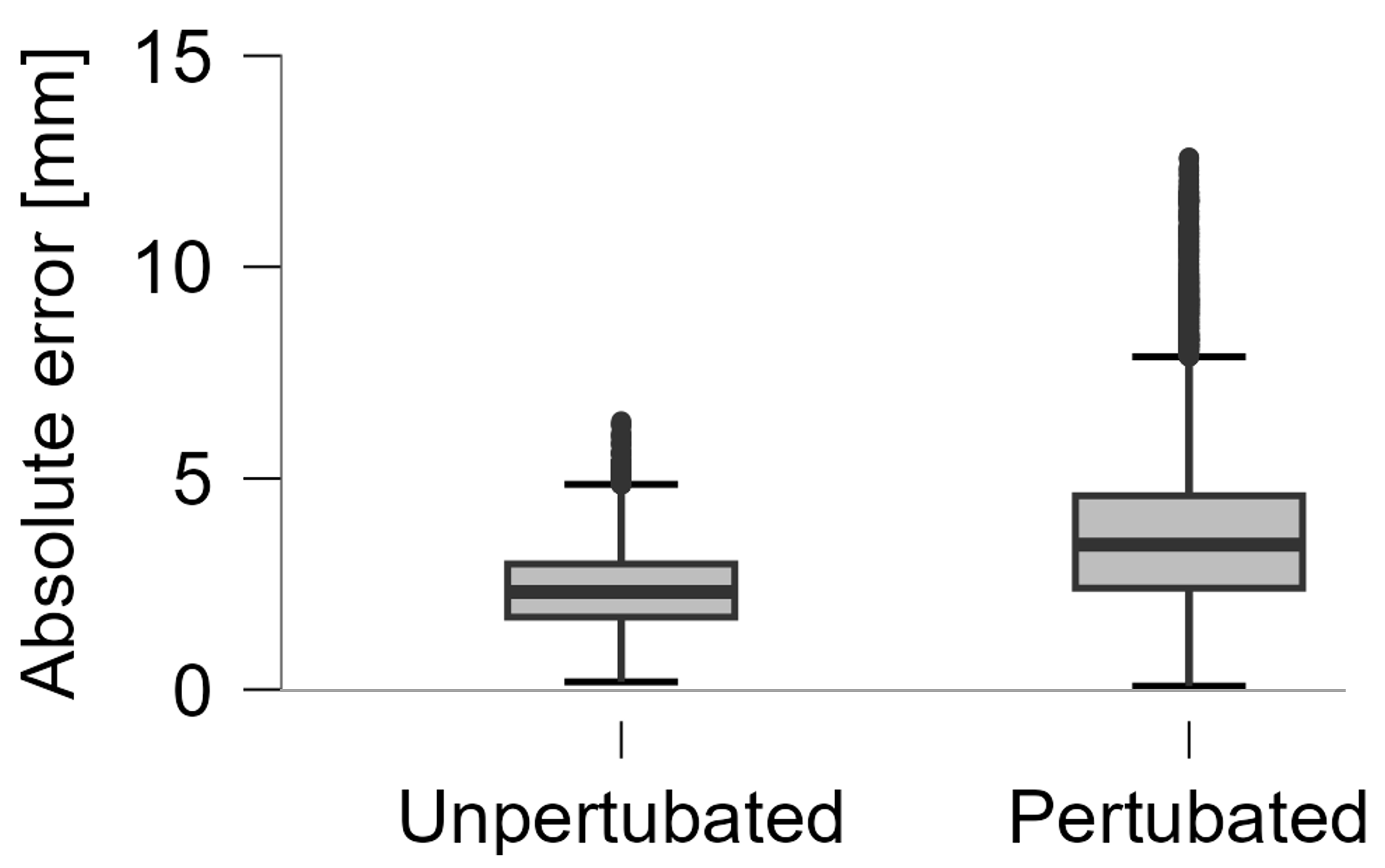}
\caption{\label{fig::BoxplotPerubatedEnvironment} Box plots of the absolute error [mm] in unperturbed vs perturbed environments at a 10 cm range of motion.}
\end{figure}

The results in Figure \ref{fig::BoxplotPerubatedEnvironment} show that environmental perturbations increase the error from 2.574 $\pm$ 1.067 mm in the unperturbed setup to 3.750 mm $\pm$ 1.996 mm in the perturbed setup. The maximum error increased significantly, from 5.584 mm to 12.584 mm. This indicates that while the system remains relatively robust under perturbation, the accuracy of the VIVE tracker is still affected by changes in the environment, particularly when objects are moved or people walk through the tracking area. Despite our concerted efforts to heavily perturb the environment, the tracker maintained a reasonable level of precision.

\subsection{Impact of the movement velocity}

The velocity limits for each displacement size varied due to different constraints, including the robot’s mechanical limitations and the dynamics of the movement patterns. This is why we choose pattern B (10 cm size of displacement) for this experiment since it allows us to have the widest range of velocities. In this section, we analyze how these velocities affect the tracking performance of the VIVE Ultimate Tracker across different movement types.

\begin{figure}[!h]
\centering
\includegraphics[width=0.45\textwidth]{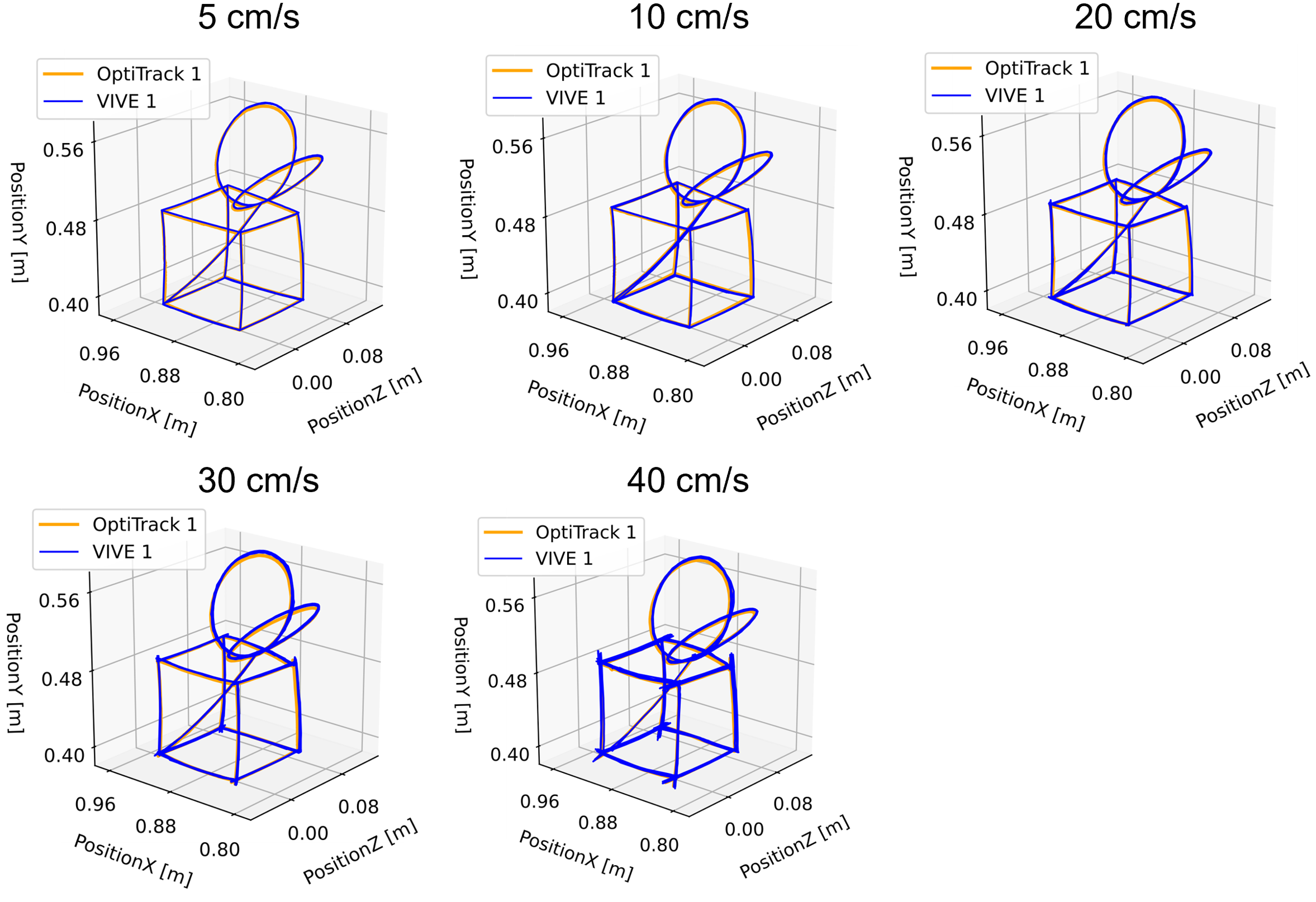}
\caption{\label{fig::diffVel_3Dplot} 3D plot of the measurements done by the OptiTrack system (orange) and the VIVE tracker (blue) at 5 different velocities}
\end{figure}

Figure \ref{fig::diffVel_3Dplot} illustrates 2-minute measurements for each end effector velocity at 120Hz, where the pattern including the linear and circular movement is repeatedly executed for each velocity. 

\paragraph{Differentiated Analysis: Cube and Circle Patterns}

Given the fundamental differences between the cube patterns, where the robot accelerates and decelerates between each corner, and the circle patterns, where the robot maintains a constant velocity, we analyzed these data sets separately. This approach allows us to gain more specific insights into how different movement dynamics affect the tracking performance of the VIVE Ultimate Tracker. In the following section, we further investigate the direct correlation between the measured velocity and the observed error by qualitatively analyzing plots that show the spatial distances between the VIVE and OptiTrack measurements - what we defined here as the absolute error - and the measured velocity by the OptiTrack system.

\paragraph{Cube Pattern - Accelerated Movements}

\begin{figure}[!h]
\centering
\includegraphics[width=0.48\textwidth]{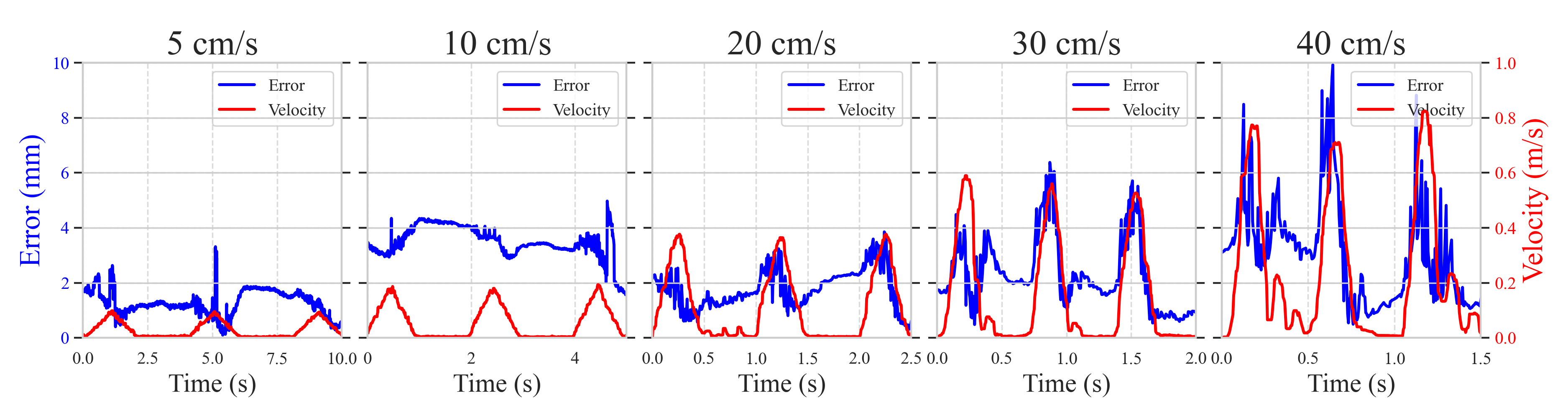}
\caption{\label{fig::Plot_Error_Velocity_vALL_cubes} Plot of the unfilterd tracking error (blue) and velocity (red) for accelerated movements at average velocities of 5 cm/s, 10 cm/s, 20 cm/s, 30 cm/s, and 40 cm/s in cube movement patterns. The velocity is measured by the OptiTrack system.}
\end{figure}

In Figure \ref{fig::Plot_Error_Velocity_vALL_cubes}, we present a subset of the data, displaying 3 linear movements out of our measurements for each of the five different average velocities. The figure plots the tracking error in the top row and the corresponding velocity, as measured by the OptiTrack system, in the bottom row.

\begin{table}[!h]
\centering
{
\begin{tabular}{lrrrrr}
& $5 cm/s$ & $10 cm/s$ & $20 cm/s$ & $30 cm/s$ & $40 cm/s$ \\
\hline
Mean & $1.491$ & $2.587$ & $1.980$ & $3.158$ & $3.073$ \\
Std. Deviation & $0.504$ & $0.812$ & $1.214$ & $2.310$ & $1.657$ \\
Maximum & $3.655$ & $7.170$ & $16.495$ & $18.144$ & $19.712$ \\
\end{tabular}
}
\caption{Summary of the measured error [mm] for different end effector velocities when performing a 10 cm range of motion movement (Pattern B, cubes only), with maximum velocities ranging from 10 cm/s to 80 cm/s, corresponding to average velocities from 5 cm/s to 40 cm/s.}
\label{tab:DiffVel_Cubes}
\end{table}

\begin{figure}[!h]
\centering
\includegraphics[width=0.35\textwidth]{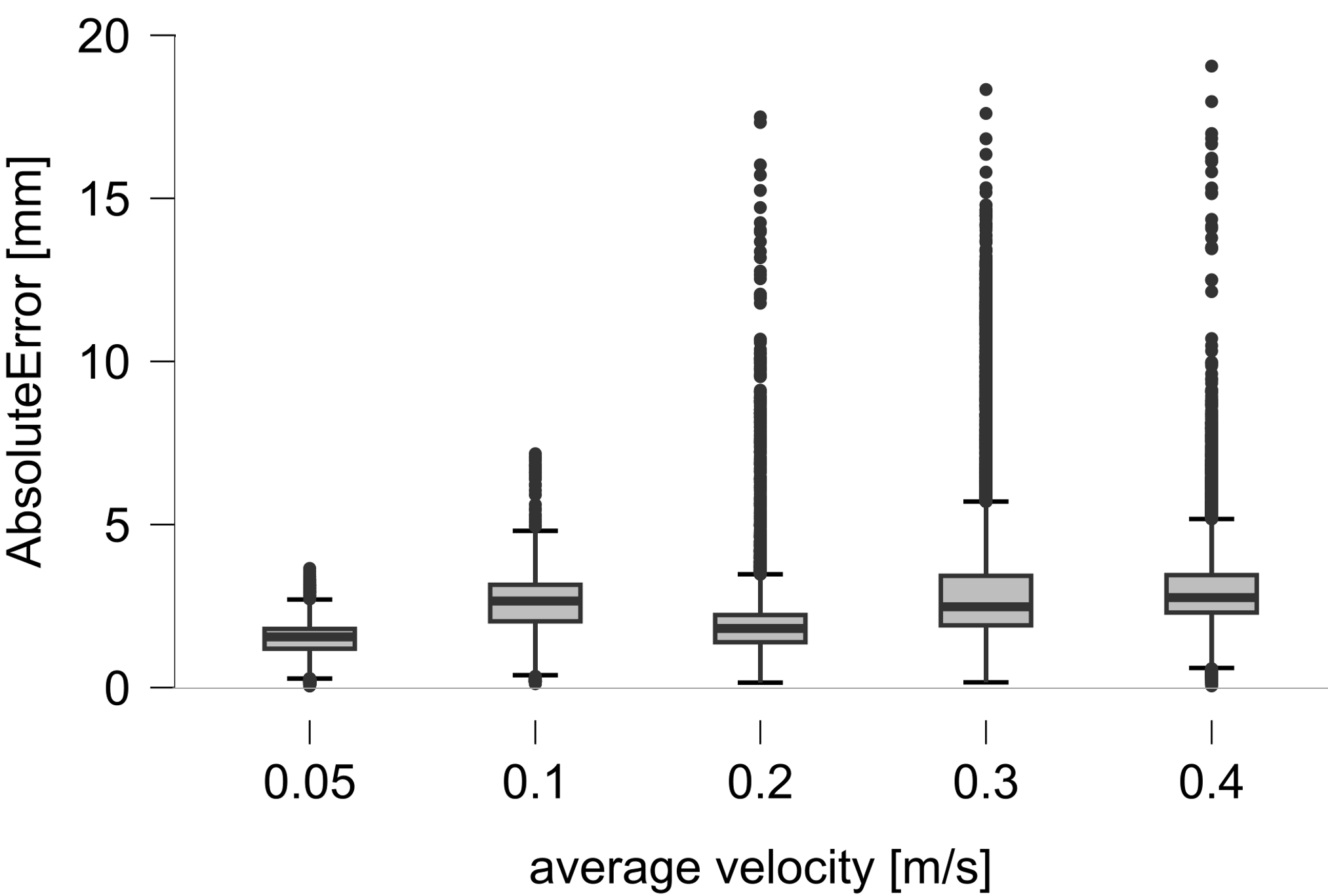}
\caption{\label{fig::BoxplotDiffVel_cubes} Box plots of the absolute error in cube movements with average velocities ranging from 5 cm/s to 40 cm/s.}
\end{figure}

The Kruskal-Wallis test indicated statistical significance for these results, with $p<0.001$, however the visual interpretation tells us a relevant correlation between the velocity and the error at around 30 cm/s as can be seen in \ref{fig::Plot_Error_Velocity_vALL_cubes} and a higher number of outliers can be observed in \ref{fig::BoxplotDiffVel_cubes}.

Notably in Figure \ref{fig::Plot_Error_Velocity_vALL_cubes}, the error reaches its peak at the highest velocity rather than at the points where the movement starts or ends. This observation suggests that velocity, rather than acceleration, is the primary factor influencing the tracking performance.

\paragraph{Circle Pattern - Constant Velocity Movements}

\begin{figure}[!h]
\centering
\includegraphics[width=0.48\textwidth]{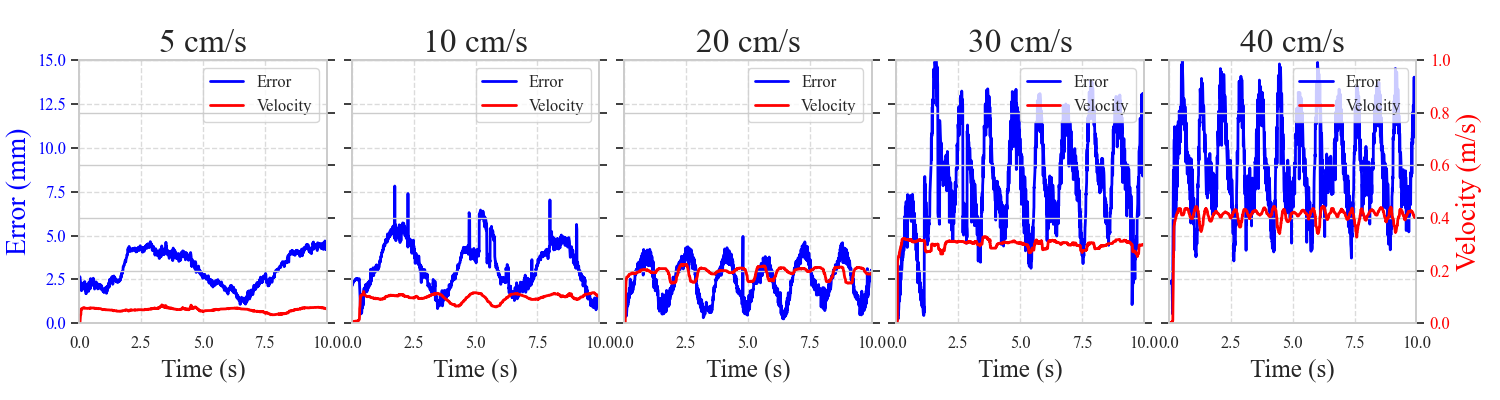}
\caption{\label{fig::velocity_xyz_t_circle} Plot of unfiltered tracking error (blue) and velocity (red) for constant velocities of 5 cm/s, 10 cm/s, 20 cm/s, 30 cm/s, and 40 cm/s in circular movement patterns. The velocity is measured by the OptiTrack system.}
\end{figure}

In Figure \ref{fig::velocity_xyz_t_circle}, the circular movement pattern shows a similar trend where the velocity significantly influences the tracking error, particularly at velocities exceeding 30 cm/s. There is a noticeable oscillation in the absolute error across all velocities, but at 30 cm/s, the amplitude of this oscillation increases markedly. Given that these oscillations occur consistently under constant velocities, with varying amplitudes and periods depending on the speed, it suggests that factors beyond velocity are contributing to the error. This observation is further explored in the following section (Section \ref{Change of displacement size}), where we analyze the effects of displacement size and proximity to the calibration center on tracking precision.

\begin{table}[!h]
	\centering
	{
		\begin{tabular}{lrrrrr}
			 & $5 cm/s$ & $10 cm/s$ & $20 cm/s$ & $30 cm/s$ & $40 cm/s$  \\
			 \hline
			Mean & $2.477$ & $3.229$ & $2.346$ & $6.687$ & $7.013$  \\
			Std. Deviation & $0.903$ & $1.423$ & $1.147$ & $3.417$ & $3.868$  \\
			Maximum & $4.672$ & $7.834$ & $6.300$ & $13.465$ & $14.880$  \\
		\end{tabular}
		\caption{Table summarizing the measured error [mm] for different velocities at a 10 cm movement size (Pattern B, circles only), with velocities ranging from 5 cm/s to 40 cm/s, corresponding.}
	    \label{tab:diffVel_circles}
	}
\end{table}

\begin{figure}[!h]
\centering
\includegraphics[width=0.35\textwidth]{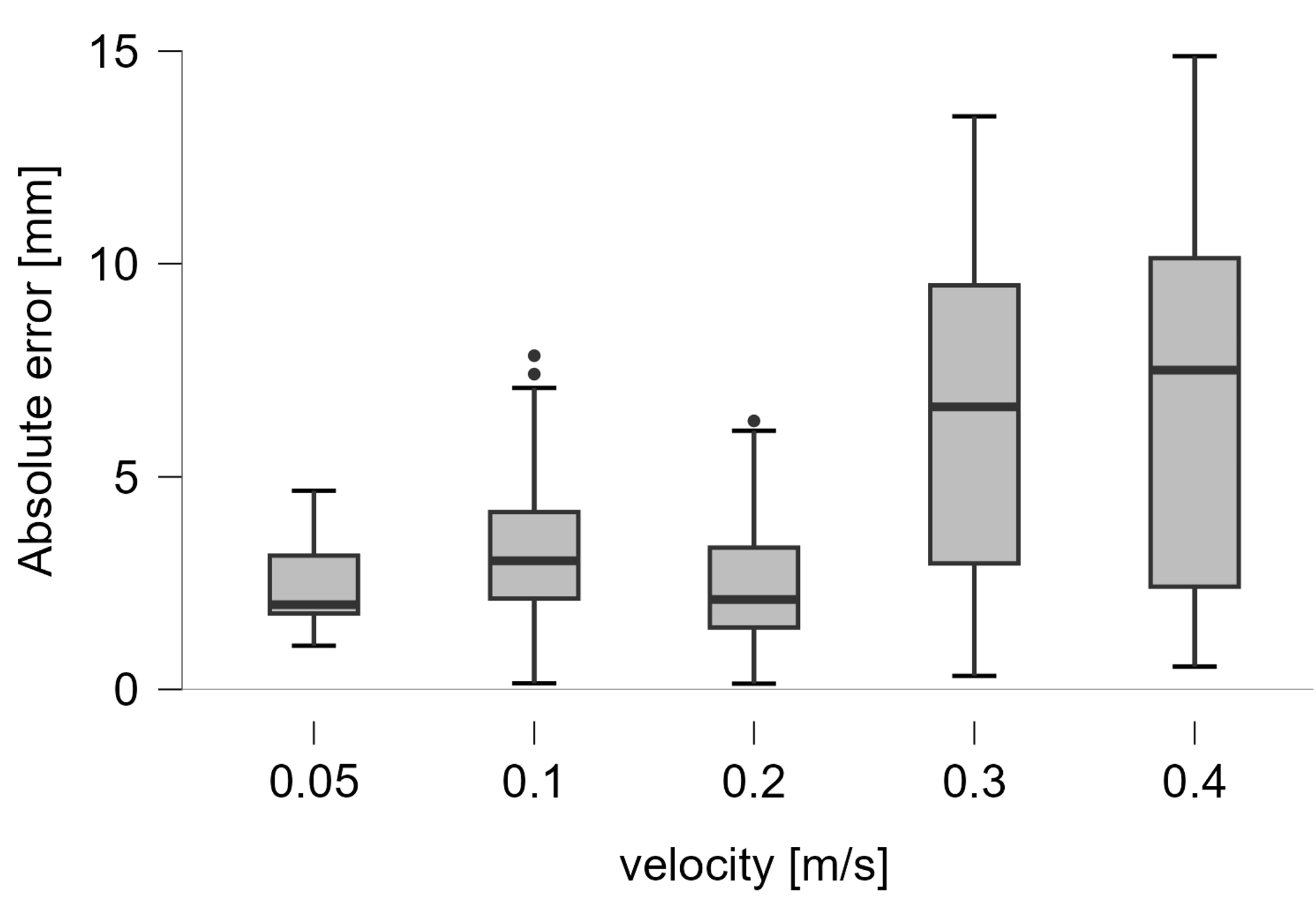}
\caption{\label{fig::BoxplotDiffVel_circles} Box plot of the error in circular movements with average velocities ranging from 5 cm/s to 40 cm/s.}
\end{figure}

Looking at both the circular and cube movement patterns, it appears that a velocity of around 30 cm/s is where a notable decline in precision begins to occur. In the circular movements, the differences in precision between 0.05 m/s and 0.2 m/s are not particularly significant; however, at 30 cm/s, there is a clear reduction in precision. This observation is further supported by the presence of more outliers in the error distributions, as illustrated in Figures \ref{fig::BoxplotDiffVel_cubes} and \ref{fig::BoxplotDiffVel_circles}.

\subsection{Impact of the displacement size}
\label{Change of displacement size}

To evaluate the impact of displacement size on tracking precision, we conducted experiments with displacement sizes ranging from pattern A-D, as illustrated in Figure \ref{fig::RobotMovements}. We maintained a constant average movement velocity of 10 cm/s to ensure that all movement patterns  could be performed. This velocity was selected as it balances the need for executing all patterns while remaining close to application realistic movement speeds, as discussed in Section \ref{Procedure}.

\begin{table}[h!]
	\centering
	{
		\begin{tabular}{lrrrr}
			 & $(A)$ & $(B)$ & $(C)$ & $(D)$  \\
			 \hline 
			Mean & $3.246$ & $2.922$ & $3.975$ & $6.757$  \\
			Std. Deviation & $0.771$ & $1.094$ & $1.512$ & $2.271$  \\
			Maximum & $17.278$ & $8.163$ & $11.953$ & $12.114$  \\
		\end{tabular}
		\caption{Summary of the measured error [mm] for different displacement sizes (Pattern A-D) at a 10 cm/s average movement velocity.}
	\label{tab:descriptiveStatistics}
	}
\end{table}

\begin{figure}[!h]
\centering
\includegraphics[width=0.3\textwidth]{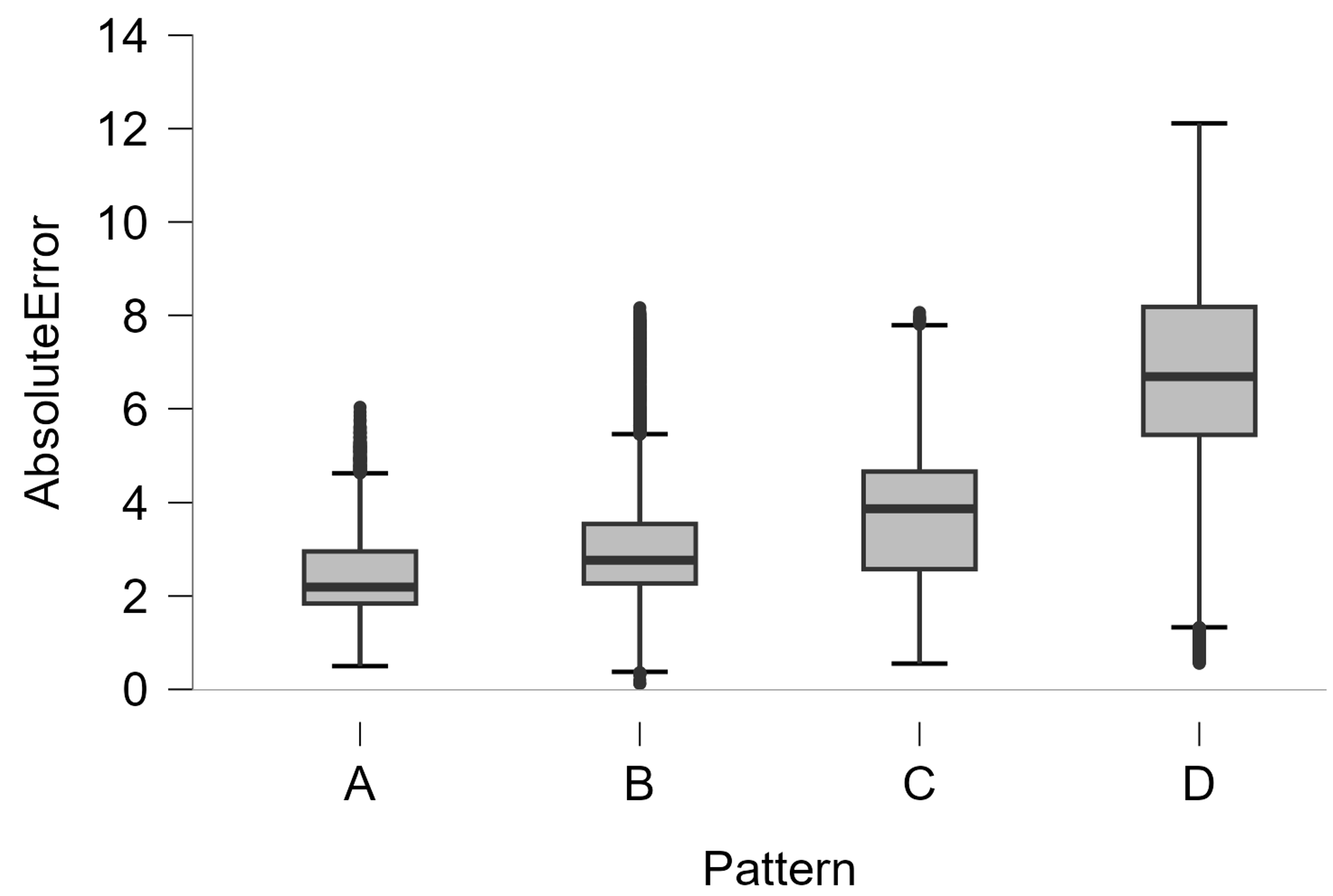}
\caption{\label{fig::Boxplot_differentMovementSizes} Box plot of the absolute error [mm] for different displacement sizes, ranging from 5 cm (Pattern A) to 80x40x20 cm (Pattern D).}
\end{figure}

\begin{figure}[!h]
\centering
\includegraphics[width=0.48\textwidth]{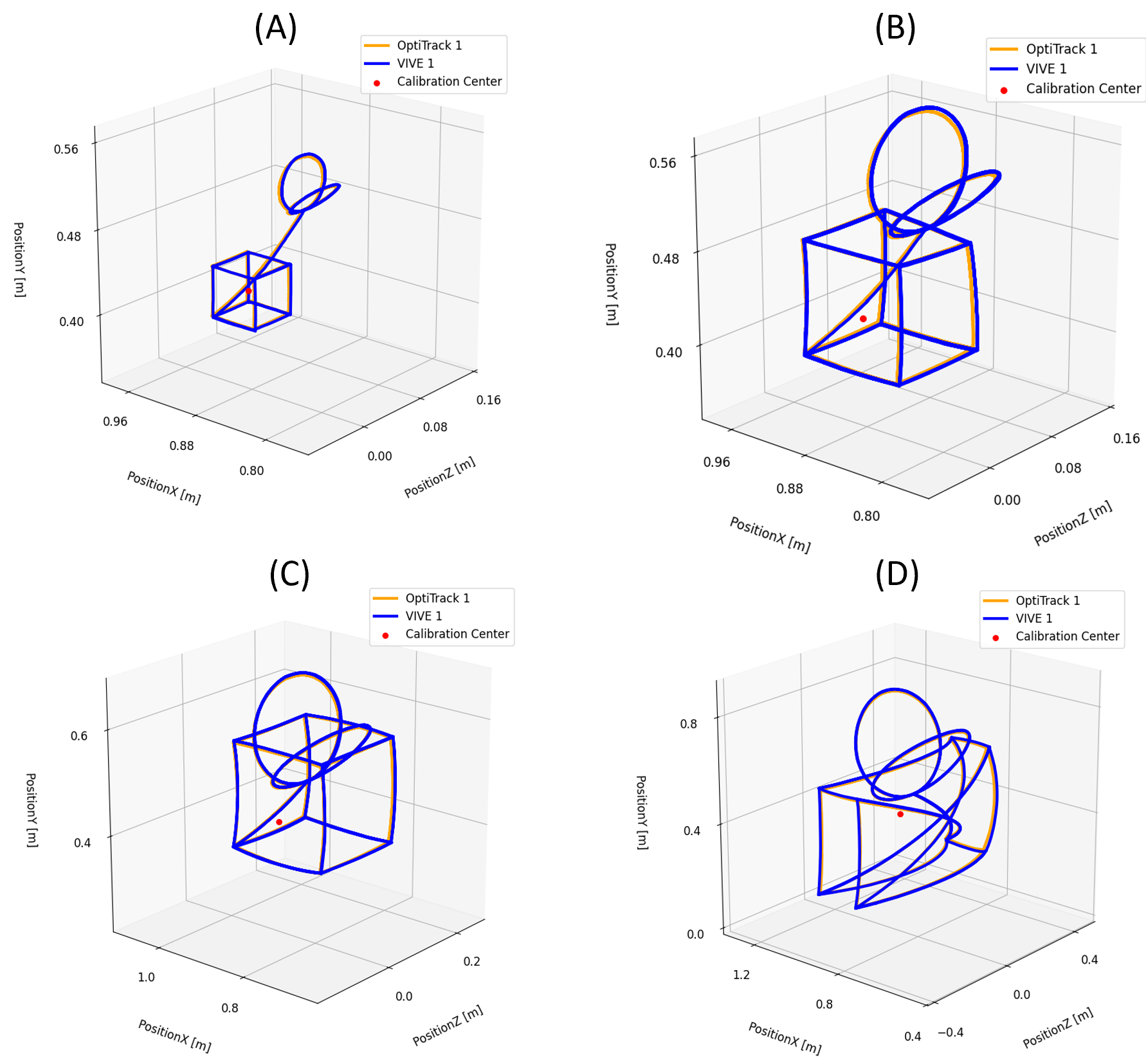}
\caption{\label{fig::3D_v01_differentS} 
3D plot comparing VIVE measurements (blue line) and OptiTrack measurements (orange line) across various displacement sizes. The calibration center is indicated by the red dot.}
\end{figure}

The data presented in Table \ref{tab:descriptiveStatistics} and the corresponding box plot (Figure \ref{fig::Boxplot_differentMovementSizes}) reveal a clear trend: larger displacement sizes lead to greater absolute error. While the correlation between displacement size and error is less pronounced for smaller displacements (e.g., 5 cm and 10 cm), it becomes increasingly significant for larger displacements, particularly for pattern D with a size of around 80 x 40 x 20 cm. The 3D plot in Figure \ref{fig::3D_v01_differentS} further illustrates the divergence between the VIVE and OptiTrack measurements as displacement size increases.

To investigate the correlation between displacement from the calibration center and tracking error, we introduced a new parameter: the distance to the calibration center. This parameter allows us to visually analyze the impact and see the direct correlation of displacement on the tracking accuracy.

\begin{figure}[!h]
\centering
\includegraphics[width=0.48\textwidth]{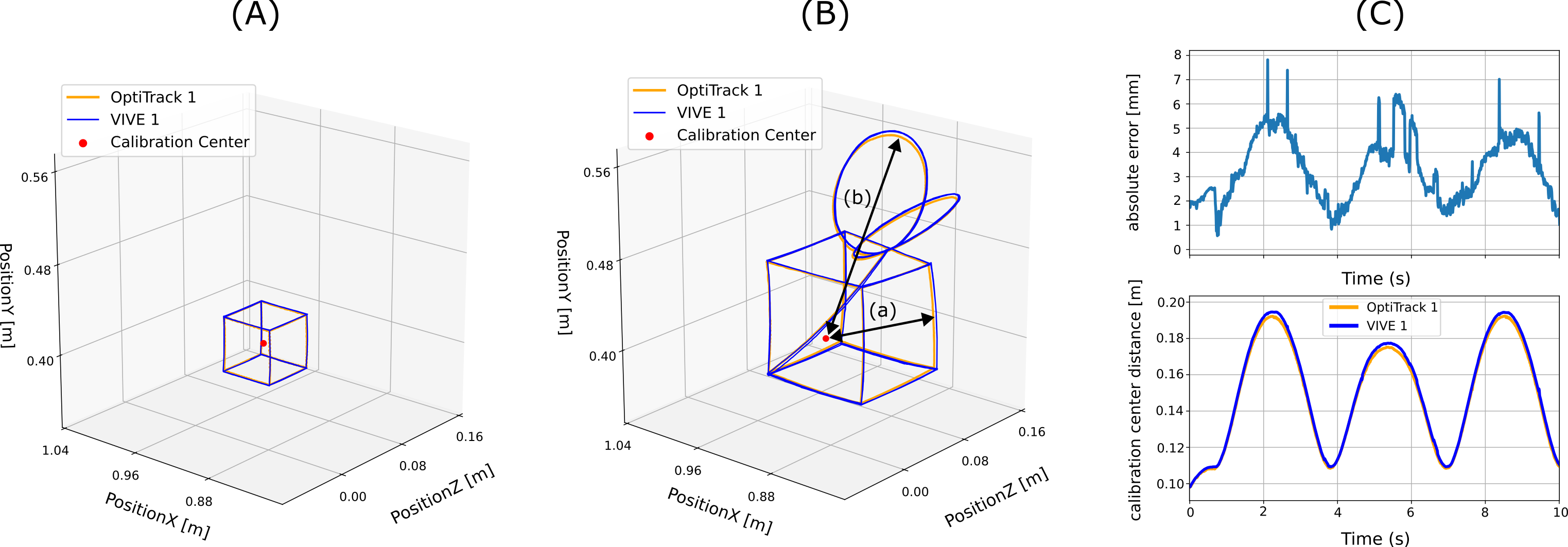}
\caption{\label{fig::ExplaininDistanceToCalibrationCenter} (A) Calibration setup with the 5 cm cube defining the calibration center (red dot). (B) Visualization of the distances to the calibration center. (C) Correlation between distance to the calibration center and unfiltered tracking error, showing increased drift between OptiTrack and VIVE measurements as distance increases.}
\end{figure}

In Figure \ref{fig::ExplaininDistanceToCalibrationCenter}, panel (A) shows the 5 cm cube movement used for initial calibration, with the red dot indicating the center of this cube, which we define as the calibration center. In panel (B), these distances are visualized for the tracking points, illustrating how we measure the displacement from the calibration center.

Panel (C) presents a section of the data from the circular movement pattern at an average velocity of 10 cm/s (Pattern B, with a 10 cm circle diameter). The plot reveals a clear correlation between the distance from the calibration center and the tracking error. Moreover, it shows how the discrepancy between the OptiTrack and VIVE measurements increases as the distance from the calibration center grows. This is also evident in the indicated distances (a) and (b) in Panel B, where the divergence between the OptiTrack and VIVE measurements is clearly visible.

To further investigate this effect, we performed additional experiments using a 20 cm displacement (Pattern C, with 10 cm/s average velocity) at various distances from the calibration center. The robot, mounted on a movable wagon, was displaced using a grid displacement pattern with varying x and y offsets from the calibration center.

\begin{figure}[!h]
\centering
\includegraphics[width=0.35\textwidth]{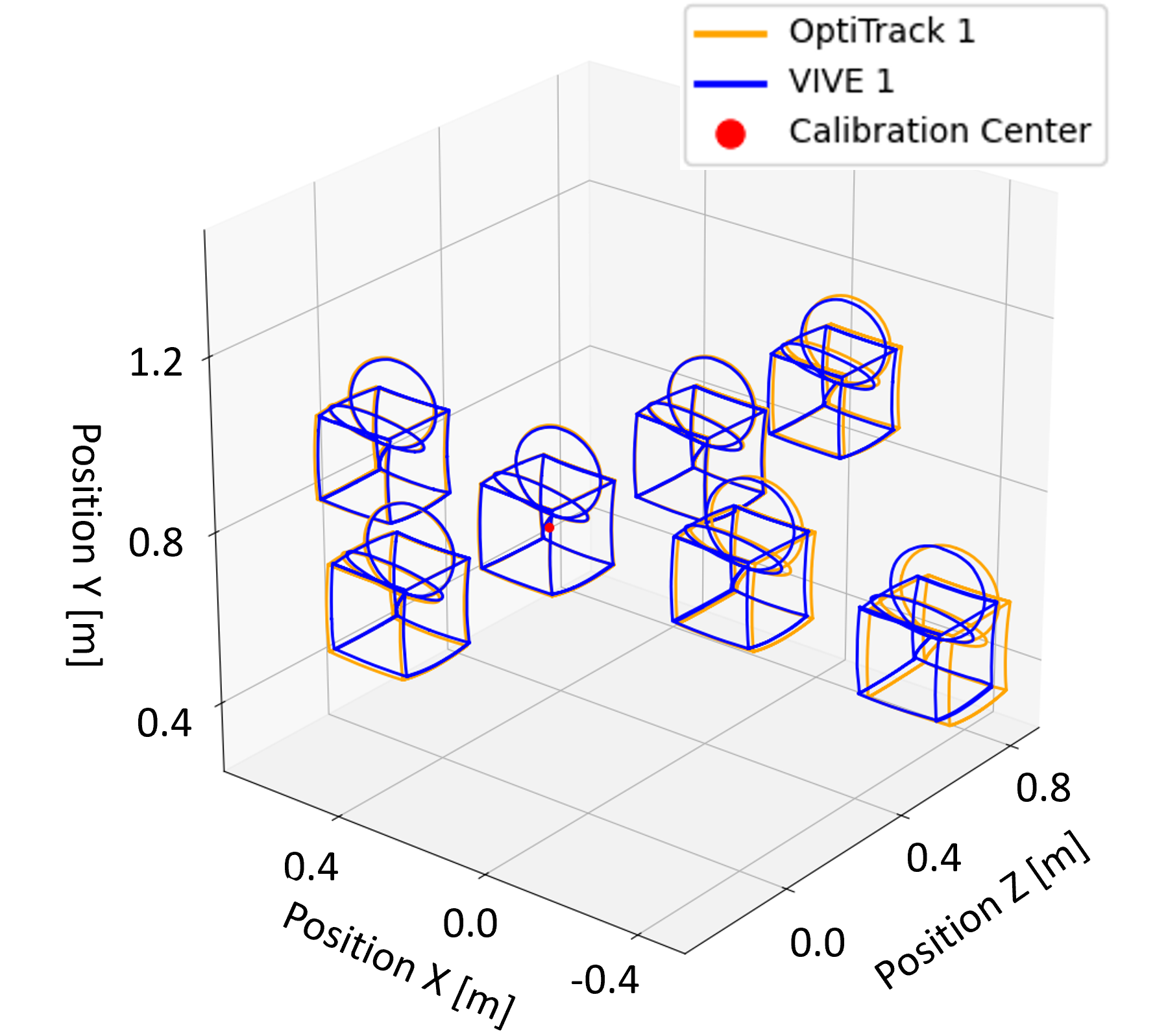}
\caption{\label{fig::3Dplot_DisplaceRobot_grid} 3D plot of the measured displacements along the x and z axes, with each data point representing the VIVE tracker's positional error relative to the calibration center.}
\end{figure}

The results, summarized in Table \ref{tab:DisplaceRobot_nonRel_grid} and illustrated in Figure \ref{fig::3Dplot_DisplaceRobot_grid}, show a significant increase in absolute error as the distance from the calibration center grows. This trend suggests that the accuracy of the VIVE tracker decreases with greater displacement from the calibration center, aligning with our earlier findings on displacement size. Interestingly, this experiment reveals an asymmetry in how the error increases across different directions. The error escalates more sharply in the x direction compared to the z direction, and within these, the error is more pronounced in the negative x and z directions than in the positive. The most extreme case occurs at a -100 mm displacement in the x direction, where the shift in position between the OptiTrack (orange) and VIVE (blue) measurements is evident in Figure \ref{fig::3Dplot_DisplaceRobot_grid}. In this instance, the scaling effect is negative, meaning that the VIVE tracker underestimates the actual movement. During our testing, we observed instances of both underestimation and overestimation by the VIVE tracker. In some cases, this scaling issue, as noted in Section \ref{CalibrationDiscussion}, was apparent during the calibration process itself, where a clear offset between the VIVE and OptiTrack measurements was visible after applying the transformation. However, in the results presented in this section, the observed error was minimal after the calibration, with a mean spatial distance between OptiTrack and VIVE measurements of only 0.8 mm, so no scaling issues were visible in the 5 cm calibration cube.

\begin{table}[h]
	\centering
	{
        \begin{tabular}{lrrrr}
        robot placement & Mean & Std. Deviation & Max \\
        \hline
        center & $3.524$ & $1.414$ & $13.9$ \\
        +50 [x] & $8.555$ & $2.524$ & $16.9$ \\
        -50 [x] & $12.963$ & $2.840$ & $21.1$ \\
        -100 [x] & $26.002$ & $3.475$ & $34.5$ \\
        +50 [z] & $5.971$ & $1.548$ & $10.5$ \\
        -50 [z] & $10.491$ & $2.101$ & $15.6$ \\
        +100 [z] & $14.226$ & $1.776$ & $20.6$ \\
        \end{tabular}
		\caption{Summary of errors [mm] for different distances from the calibration center at a 10 cm/s average movement velocity and 20 cm displacement (Pattern C).}
	\label{tab:DisplaceRobot_nonRel_grid}
	}
\end{table}

\begin{figure}[!h]
    \centering
    \begin{subfigure}[b]{0.45\textwidth}
        \includegraphics[width=\textwidth]{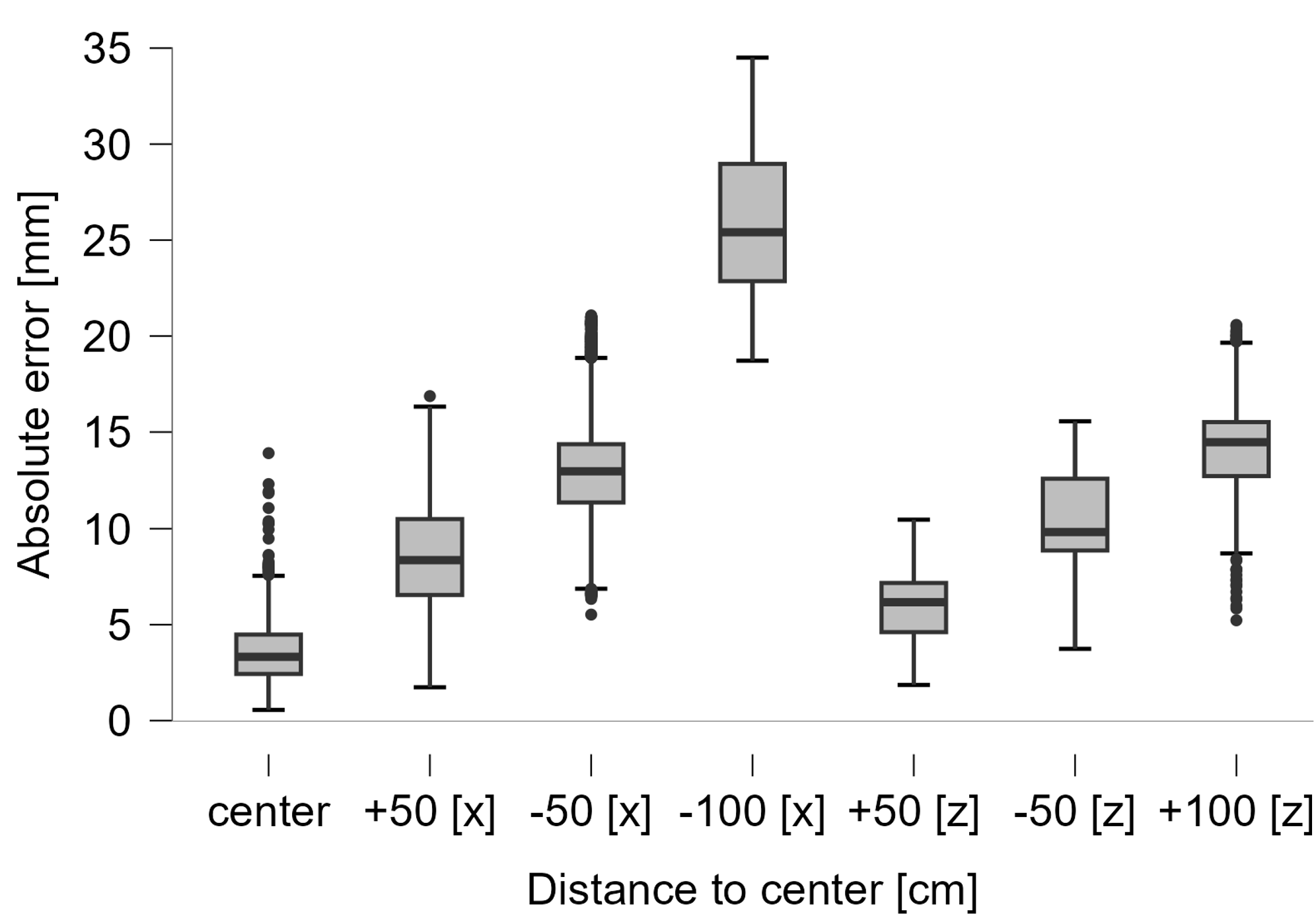}
        \caption{Box plots of the absolute error at different distances from the calibration center along the x and z axes.}
        \label{fig::box_plot_moveRobot}
    \end{subfigure}
    \hfill
    \begin{subfigure}[b]{0.45\textwidth}
        \includegraphics[width=\textwidth]{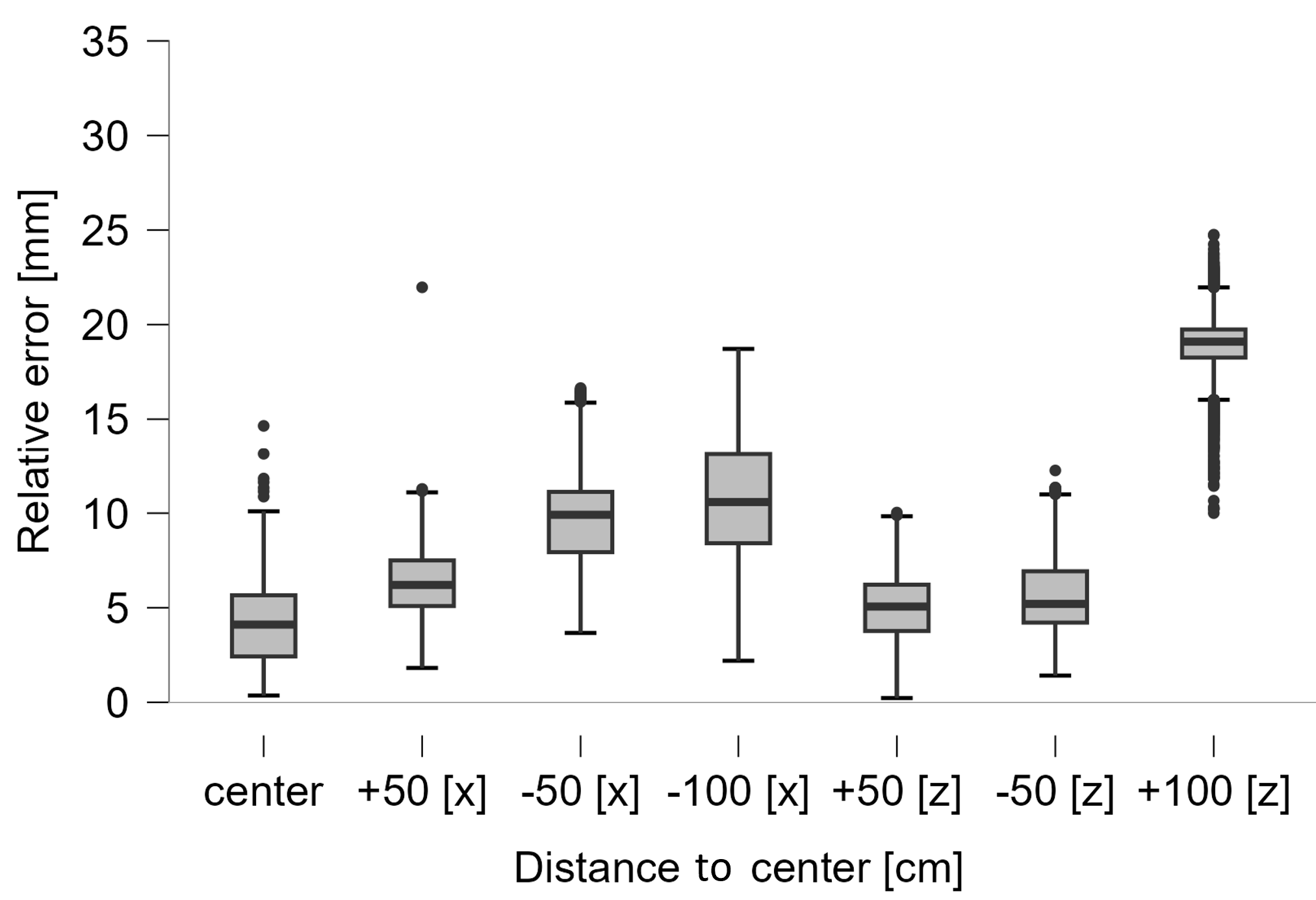}
        \caption{Box plots of the error at different distances from the calibration center along the x and z axes measured relatively to the reference tracker.}
        \label{fig::box_plot_moveRobot_relative}
    \end{subfigure}
    \caption{Box plots of the absolute error at different distances from the calibration center along the x and z axes comparing the measuring the position globally with measuring it relatively to the reference tracker.}
    \label{fig::combined_plots_robotDisplacement}
\end{figure}

Given that the VIVE tracking system is primarily designed for relative tracking in conjunction with a VR headset, we hypothesized that measuring relative movements, rather than absolute global displacements, might yield more accurate results. To test this hypothesis, we utilized the second VIVE tracker of our experimental setup. This second tracker was attached to the movable wagon, allowing us to measure displacements relative to this tracker instead of to a fixed global point.

The results, summarized in Figure \ref{fig::box_plot_moveRobot_relative}, show a noticeable reduction in error for relative displacements compared to global measurements. This finding suggests that the VIVE system's design for relative tracking does indeed perform better when used in such a manner. However, it is important to note that in some cases, such as at +100 cm in the z direction, the error was not reduced. This indicates that while relative tracking can improve accuracy, it may still encounter limitations depending on the specific movement or direction of displacement.

To conclude, our analyses identified displacement size as a key factor influencing the precision of the VIVE Ultimate Tracker. Larger displacements and increased distances from the calibration center were consistently linked to higher positional errors. This trend highlights the importance of maintaining proximity to the calibration center to ensure optimal tracking accuracy, as moving further away tends to introduce greater error, likely due to inherent geometric distortions or scaling issues within the VIVE’s tracking system.

\newpage
\subsection{Evaluation of performance in capturing human movement}
\paragraph{Pick and Place}

\begin{figure}[!h]
\centering
\includegraphics[width=0.45\textwidth]{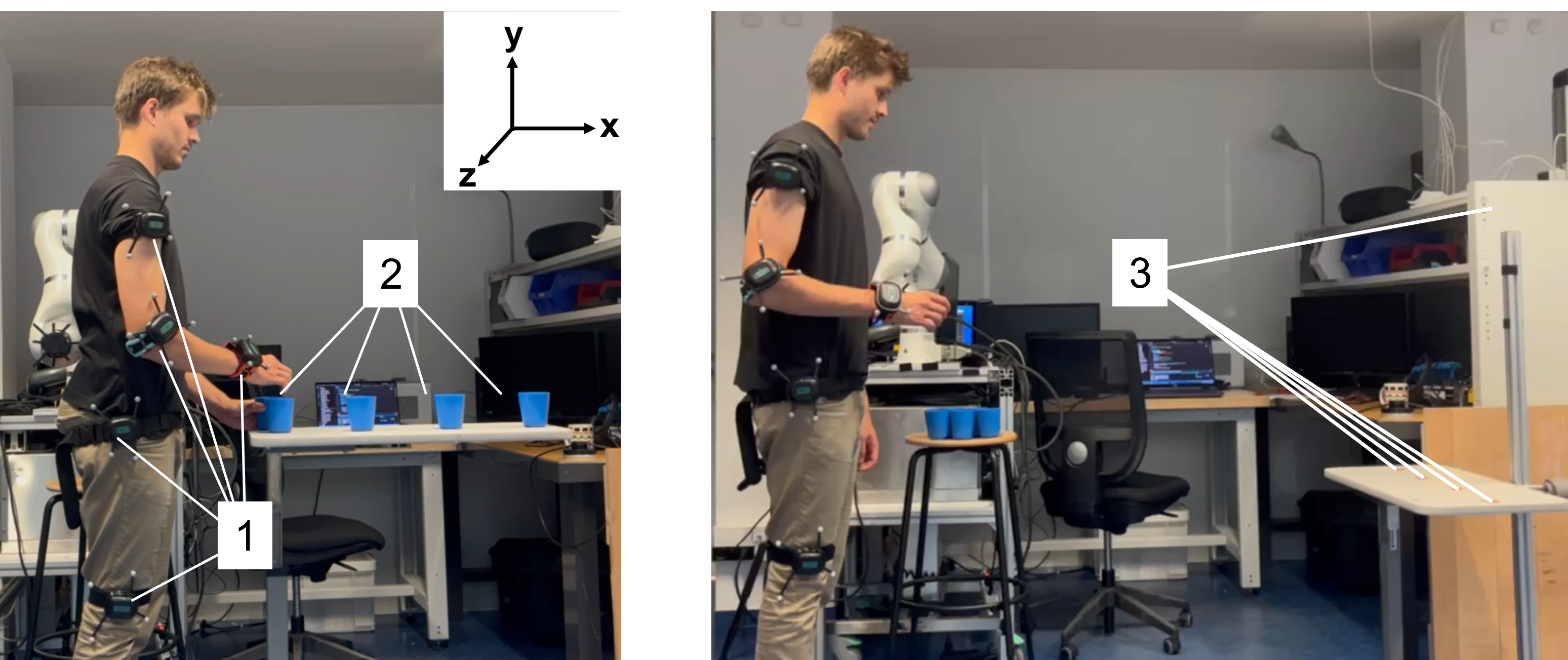}
\caption{\label{fig::PickandPlaceSetup} Experimental setup for two pick-and-place tasks using five attached trackers with each one OptiTrack marker set (1): knee, hip, shoulder, elbow, and wrist. Left panel: task performed while standing still. Right panel: task performed with walking. The goal destinations are marked by 2 in the left panel and by 3 in the right panel.}
\end{figure}

In the first experiment, we assessed the performance of multiple trackers during a straightforward pick-and-place task. The subject was instructed to pick up cups and place them at varying distances while standing, with a range of motion (ROM) in the static case of approximately 0.7 meters along the x-axis, 0.15 meters along the y-axis, and 0.1 meters along the z-axis. In the second experiment, we increased the task complexity by positioning the goal destinations at different distances and heights, requiring the subject to walk forward and backward to reach the targets. This introduced a broader range of motion, with the dynamic case ROM extending to 1.5 meters along the x-axis, 0.7 meters along the y-axis, and 0.7 meters along the z-axis. Figure \ref{fig::PickandPlaceSetup} illustrates the two experimental setups. Trackers were affixed to key points on the subject's body, including the wrist, elbow, shoulder, hip, and knee.

\begin{figure}[!h]
\centering
\includegraphics[width=0.45\textwidth]{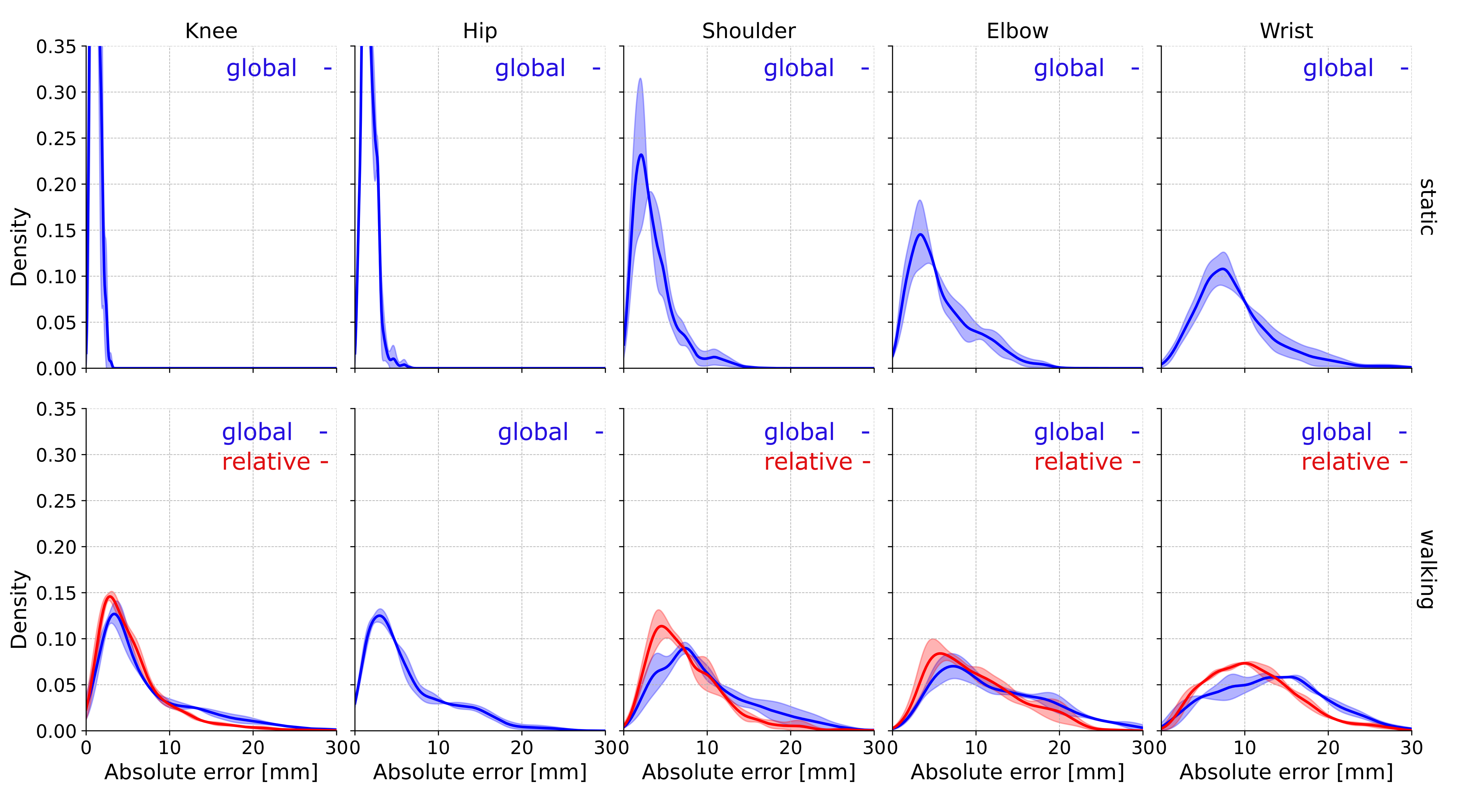}
\caption{\label{fig::PickAndPlaceDistributionPlots} Density plots illustrating the distribution of absolute errors across three trials for each body part (knee, hip, shoulder, elbow, and wrist) during the pick-and-place tasks. The top row depicts the static condition, where the subject remained stationary, while the bottom row represents the walking condition, where the subject moved back and forth. In the bottom row, the errors calculated relative to the hip tracker are shown in red, as a comparison with the global error in blue.}
\end{figure}

Figure \ref{fig::PickAndPlaceDistributionPlots} illustrates the distribution of absolute errors across three trials for each tracker (knee, hip, shoulder, elbow, and wrist) during both static and walking pick-and-place tasks. In the static experiment, the knee and hip trackers show very narrow error distributions, with mean errors of 1.62 mm $\pm$ 0.80 mm and 1.05 mm $\pm$ 0.49 mm, respectively. This low error can be attributed to the minimal movement at these joints during the static task, particularly as the calibration was done at the hip level where the subject stood.

The wrist tracker exhibits the highest error, with a mean of 8.93 mm $\pm$ 5.01 mm, followed by the elbow (5.52 mm $\pm$ 3.64 mm) and shoulder (3.60 mm $\pm$ 2.62 mm). These results suggest that the further the tracker moves, the greater the error.

In the walking experiment, the error distribution shifts, with the mean errors increasing across all trackers. The hip and knee errors rise to 6.65 mm $\pm$ 5.07 mm and 7.49 mm $\pm$ 6.00 mm, respectively. The upper body trackers continue to show greater errors, with the wrist reaching 13.68 mm $\pm$ 6.37 mm, the elbow 12.19 mm $\pm$ 6.46 mm, and the shoulder 9.98 mm $\pm$ 5.70 mm. This increase in error during the walking task highlights the impact of more dynamic movements on tracking accuracy, particularly for larger displacements.

\paragraph{Fencing}
In our final experiment, inspired by the pioneering work of Etienne-Jules Marey from 1890 \cite{Marey1890}, we challenged an elite fencer to perform repetitive lunges at high speed with a considerable range of motion. This experiment aims to push the boundaries of our motion tracking system by capturing a movement that is both highly dynamic and involves a wide range of motion. We again tracked the hip, knee, shoulder, wrist, and elbow. This experiment is designed to test the limits of the VIVE tracker, especially after our previous findings identified displacement size and velocity as the primary challenges in maintaining accurate motion tracking.

\begin{figure}[!h]
\centering
\includegraphics[width=0.45\textwidth]{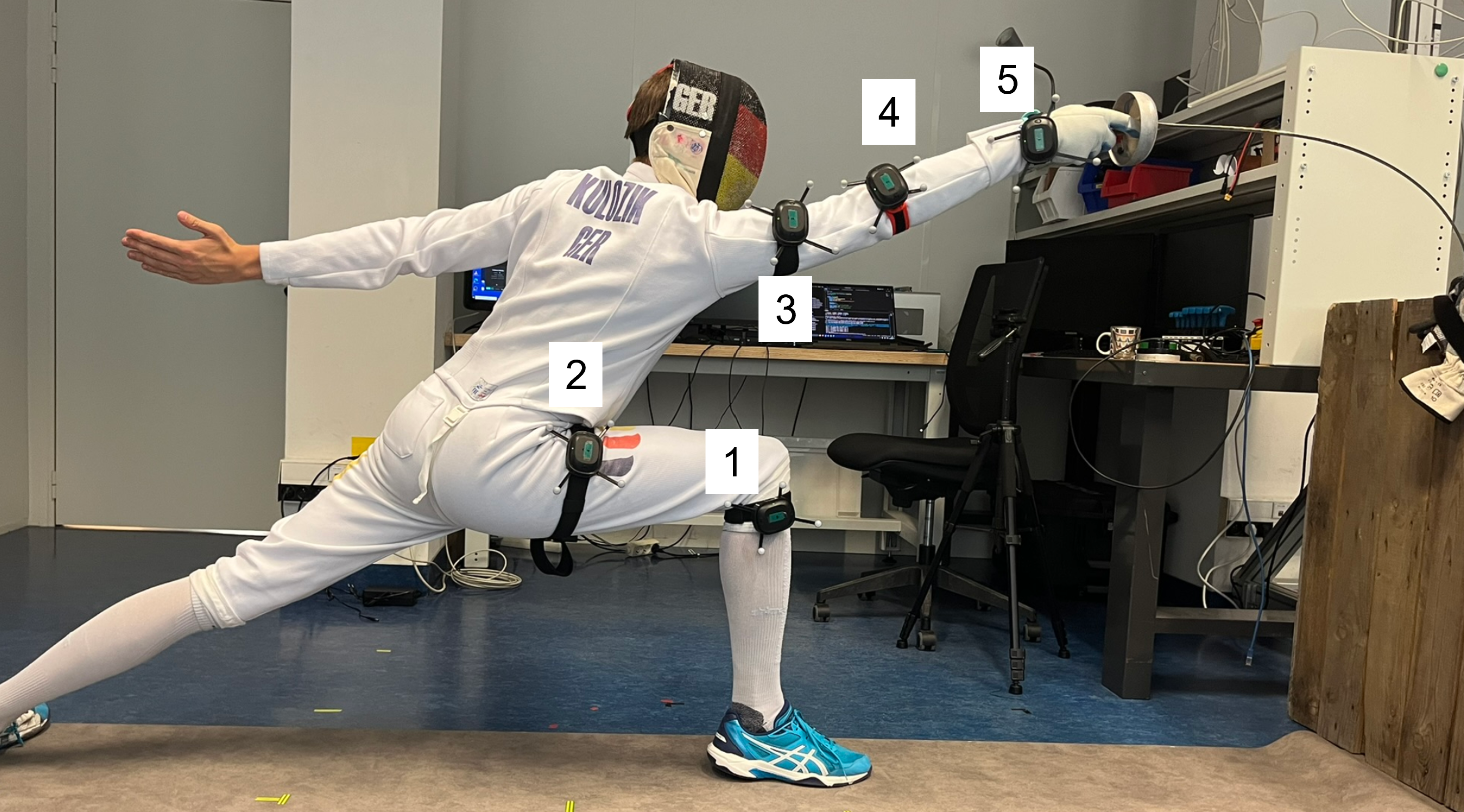}
\caption{\label{fig::FencingSetup} Experimental setup for the fencing task with trackers attached to the knee, hip, shoulder, elbow, and wrist.}
\end{figure}

\begin{table}[h]
\centering
{
\begin{tabular}{lrrrr}
	 & Mean & SD & Max. Vel [m/s] & RoM [m]\\
	 \hline
	Knee & $6.624$ & $5.557$ & $2.78$ & $\approx 1.1$\\
	Hip & $9.208$ & $6.766$ & $2.25$ & $\approx 1.0$ \\
	Shoulder & $12.202$ & $6.869$ & $4.72$ & $\approx 1.4$ \\
	Elbow & $12.309$ & $8.728$ & $6.3$ & $\approx 1.5$\\
	Wrist & $17.516$ & $10.725$ & $9.34$ & $\approx 1.6$\\
\end{tabular}
}
\caption{Measured tracking error [mm] and range of motion (ROM) [m] and maximum velocity [m/s] for each body part during the fencing experiment.}
\label{tab:errorFencing}
\end{table}

\begin{figure}[!h]
\centering
\includegraphics[width=0.45\textwidth]{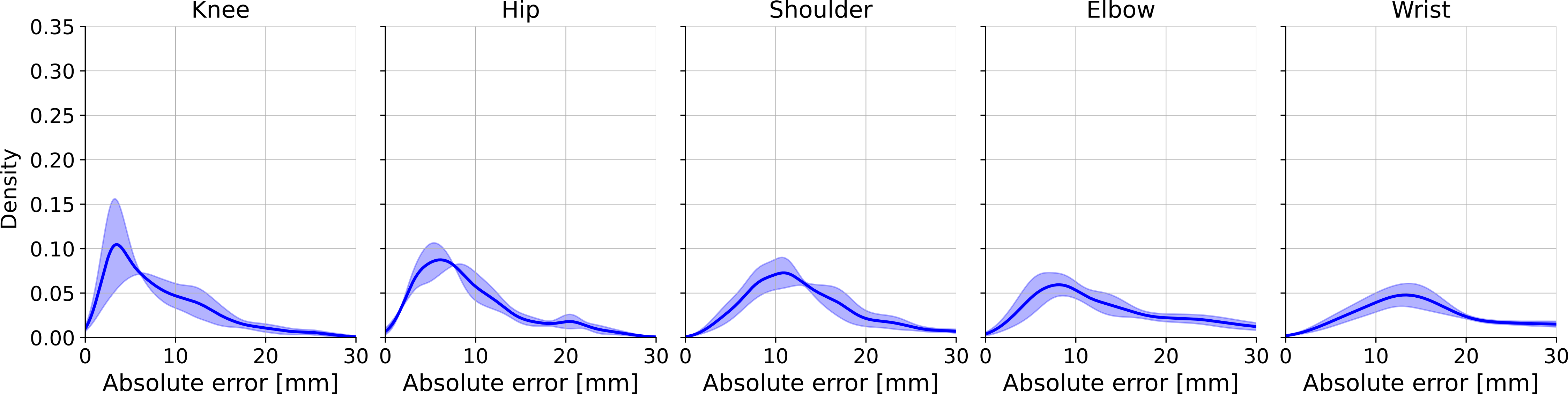}
\caption{\label{fig::DensityPlotFencing} Density plots over three trials (each trial consisting of eight lunges) showcasing the distribution of tracking errors across different body parts during the fencing task.}
\end{figure}

The results, summarized in Table \ref{tab:errorFencing} and visualized in Figure \ref{fig::DensityPlotFencing}, indicate that the tracking error increases significantly with the dynamic nature of the motion. The mean absolute errors were highest for the wrist (17.52 mm) and elbow (12.31 mm), reflecting the larger range of motion and higher velocity at these points during the lunge. The hip and knee, which have less motion relative to the body’s center, exhibited lower errors, at 9.21 mm and 6.62 mm respectively. The shoulder error was intermediate at 12.20 mm. The standard deviations are also indicative of the variability in tracking precision, with the wrist showing the greatest variability (10.73 mm), followed by the elbow (8.73 mm).

These results emphasize the challenges of tracking fast, dynamic movements with a wide range of motion, particularly at extremities like the wrist. The VIVE tracker, while relatively accurate for less dynamic movements, shows its limitations in such demanding conditions.

\section{Discussion}

The HTC VIVE Ultimate Tracker offers a compelling, cost-effective, and easy-to-use motion tracking solution, especially in situations where traditional systems like OptiTrack are impractical due to their complexity, cost, or the need for sufficient space to position cameras at a proper distance from the objects being tracked. The VIVE Tracker, leveraging inside-out tracking with stereo cameras, offers a streamlined setup free of external cameras, which greatly simplifies deployment in various environments. On average, the HTC VIVE Ultimate Tracker achieved a precision of 4.98 mm $\pm$ 4 mm across various conditions.
In best case scenario the precision was measured to be 2.59 mm $\pm$ 0.81 mm. The availability of the software provided by us, both in C++ and Python \cite{MyRepo}, further enhances its accessibility for researchers and developers who require a motion tracking solution without the sub-millimeter precision of high-end systems.

Our study highlights the utility of the VIVE Tracker across various conditions, though it does reveal some design limitations. The tracker demonstrated good precision in scenarios such as human motion during activities of daily living (ADL), but it reached its limits with highly dynamic gestures, such as capturing the high-speed movements of a fencer. This performance suggests that while the VIVE Tracker is suitable for many general motion tracking applications, it may struggle with tasks requiring sub-millimeter accuracy.

However, several factors affect the precision of the VIVE Tracker. Notably, our experiments identified scaling issues, particularly when the tracker was displaced significantly from the calibration center. This scaling problem likely arises from the system's design for relative motion tracking in a local frame, rather than global displacement tracking. Users must be vigilant about these scaling issues; before commencing any data collection, it is advisable to verify the scaling by moving the tracker next to a known scale. If discrepancies are observed, users should either correct the data manually by applying a scaling factor or re-calibrate the system. In our experiments, re-calibration was necessary whenever scaling issues were visibly off during the initial calibration cube setup.

The study also underscored the significant impact of environmental factors on the accuracy of the VIVE Tracker. Both low lighting conditions and environmental disturbances, such as moving objects or people passing through the tracker’s field of view, were found to measurably affect precision. Although the tracker maintained reasonable accuracy under these perturbed conditions, the errors notably increased, especially in low-light scenarios. This limitation suggests that the VIVE Tracker is best suited for controlled environments where lighting can be optimized, even if the environment is not entirely static. Furthermore, using the sensor outdoors appears not yet feasible due to the lack of nearby reference points and significant variations in natural sunlight.

The tracker’s performance was also evaluated in terms of movement velocity and displacement size. Our findings indicate a significant increase in error as either the velocity or the displacement size increases. This effect was particularly evident in dynamic tasks involving significant limb movements, such as in our fencing experiment, where the tracker struggled to maintain accuracy at the wrist and elbow—the extremities furthest from the body’s center.

Furthermore, the study brings attention to the lack of detailed technical specifications and hardware information for the VIVE Tracker, a gap likely due to its focus on the consumer and gaming markets rather than industrial applications. This lack of information could pose challenges for researchers who require precise data on the tracker’s performance capabilities, such as latency and sampling rate, for academic research and other rigorous applications. While this study focused on evaluating the position tracking performance, the tracker’s 6 DoF capability suggests that a follow-up study assessing its orientation tracking performance could be valuable for future research.

Despite these limitations, the VIVE Ultimate Tracker remains an excellent tool for scenarios where the need for precision is not excessively stringent or where relative motion data is more important than absolute positioning. Its ease of setup, combined with the software we provide and tutorials, makes it an attractive option for many researchers and practitioners who require millimeter-level accuracy without the overhead of more complex systems.

The significance of this study lies in its potential to offer a simpler, low-cost alternative to traditional motion capture systems, which often require expensive equipment and complex setups. By examining the conditions under which the VIVE Ultimate Tracker can achieve reasonable accuracy, we aim to facilitate its adoption in robotics and other fields that demand cost-effective and flexible motion capture solutions. This study also contributes to the ongoing discourse on the feasibility of using consumer-grade technologies for professional applications, providing insights that could guide future research and development.

Lastly, it is important to note that the VIVE Tracker's software and firmware are subject to continuous updates, which could impact the validity of the findings presented here. Users should remain aware of any updates that could alter the tracker’s performance characteristics and adjust their experimental protocols accordingly.

\section{Conclusion}

The HTC VIVE Ultimate Tracker proves to be a versatile and cost-effective alternative to traditional motion capture systems, particularly in applications where ease of use and flexibility are paramount. While it does not achieve the sub-millimeter precision of high-end optical motion capture systems with external cameras and reflective markers like OptiTrack, it provides sufficient accuracy for a wide range of motion tracking scenarios. The tracker’s performance is influenced by factors such as lighting conditions, environmental changes, velocity, and displacement size, with scaling issues presenting the most significant challenge. Despite these limitations, the VIVE Tracker's affordability, ease of setup, and adaptability make it a valuable tool for researchers and developers working in fields where high precision is not the primary requirement. Users must, however, remain vigilant about potential scaling errors and ensure proper calibration to maximize the tracker’s accuracy. As the software and hardware continue to evolve, the VIVE Ultimate Tracker is likely to become an increasingly reliable tool for motion tracking applications outside the realm of virtual reality.

\bibliographystyle{IEEEtran}
\bibliography{sample}

\newpage
\appendix
\section{Appendix} \label{Methods}

\subsection{Data Capturing}

The data were acquired from the OptiTrack system and transmitted via NatNet to a Python script, followed by OSC transmission to a laptop which was connected to the VIVE tracker via Bluetooth. Python (version 3.11.9) was utilized for data collection, statistical analyses, and visualization, alongside JASP (version 0.18.3.0).

\subsection{Registration method} \label{Registration Method}
The registration method implemented in the software aims to determine the optimal spatial transformation between two sets of three-dimensional points. These point sets represent positions captured in different coordinate systems, specifically those from a Vive tracking system and an OptiTrack system.

\subsubsection{Mathematical Formulation}
The registration process begins by calculating the centroids of both point sets. Let $\mathbf{A}$ ([3xn] with xyz and n number of captured points) being the  represent the matrix of points from the Vive system and $\mathbf{B}$ from the OptiTrack system. The centroids $\mathbf{c}_A$ and $\mathbf{c}_B$ are computed as the mean of the respective points:

\begin{equation}
\mathbf{c}_A = \frac{1}{n} \sum_{i=1}^n \mathbf{A}_i, \quad \mathbf{c}_B = \frac{1}{n} \sum_{i=1}^n \mathbf{B}_i
\end{equation}

where $n$ is the number of points in each set. The point sets are then centered by subtracting their respective centroids:

\begin{equation}
\mathbf{A}' = \mathbf{A} - \mathbf{c}_A, \quad \mathbf{B}' = \mathbf{B} - \mathbf{c}_B
\end{equation}

The optimal rotation matrix $\mathbf{R}$ that aligns $\mathbf{A}'$ to $\mathbf{B}'$ is found using Singular Value Decomposition (SVD). The SVD of the matrix product $\mathbf{A}' \mathbf{B}'^\top$ is computed as follows:

\begin{equation}
\mathbf{U}, \mathbf{S}, \mathbf{V}^\top = \text{SVD}(\mathbf{A}' \mathbf{B}'^\top)
\end{equation}

The rotation matrix $\mathbf{R}$ is then given by:

\begin{equation}
\mathbf{R} = \mathbf{V} \mathbf{U}^\top
\end{equation}

To ensure a proper rotation (i.e., a right-handed coordinate system), the determinant of $\mathbf{R}$ is checked. If $\det(\mathbf{R}) < 0$, the last column of $\mathbf{V}$ is negated before computing $\mathbf{R}$.

\subsubsection{Translation Vector}
With the rotation matrix $\mathbf{R}$ established, the translation vector $\mathbf{t}$ aligning the original point sets is computed as:

\begin{equation}
\mathbf{t} = \mathbf{c}_B - \mathbf{R} \mathbf{c}_A
\end{equation}

This vector represents the shift required, after rotation, to align the centroids of the two point sets.

\subsubsection{Application}
The derived transformation, consisting of $\mathbf{R}$ and $\mathbf{t}$, is applied to the data points from the Vive system to register them into the coordinate system of the OptiTrack setup. This approach ensures that both tracking systems are spatially aligned and both data points are measured in the same frame (in our case the OptiTrack base frame).

The transformation can be solved by the equation:
\begin{equation}
{}^{\text{Opti}}\mathbf{p}_{\text{Vive}} = {}^{\text{Opti}}\mathbf{R}_{\text{Vive}} \cdot \mathbf{p}_{\text{Vive}} + \mathbf{t}
\end{equation}

\subsection{Additional Data}
\subsubsection{Repeatability of the Measurements}

To ensure the reliability and repeatability of our measurements, and to streamline data analysis by reducing the need to demonstrate repeatability for each dataset, we conducted four repeated trials of the high-velocity (30 cm/s) movement in pattern B. The absolute error was recorded for each trial, and the data were analyzed using both visual inspection (through density and box plots) and statistical tests.

\begin{figure}[!h]
    \centering
    \begin{subfigure}[b]{0.45\textwidth}
        \includegraphics[width=\textwidth]{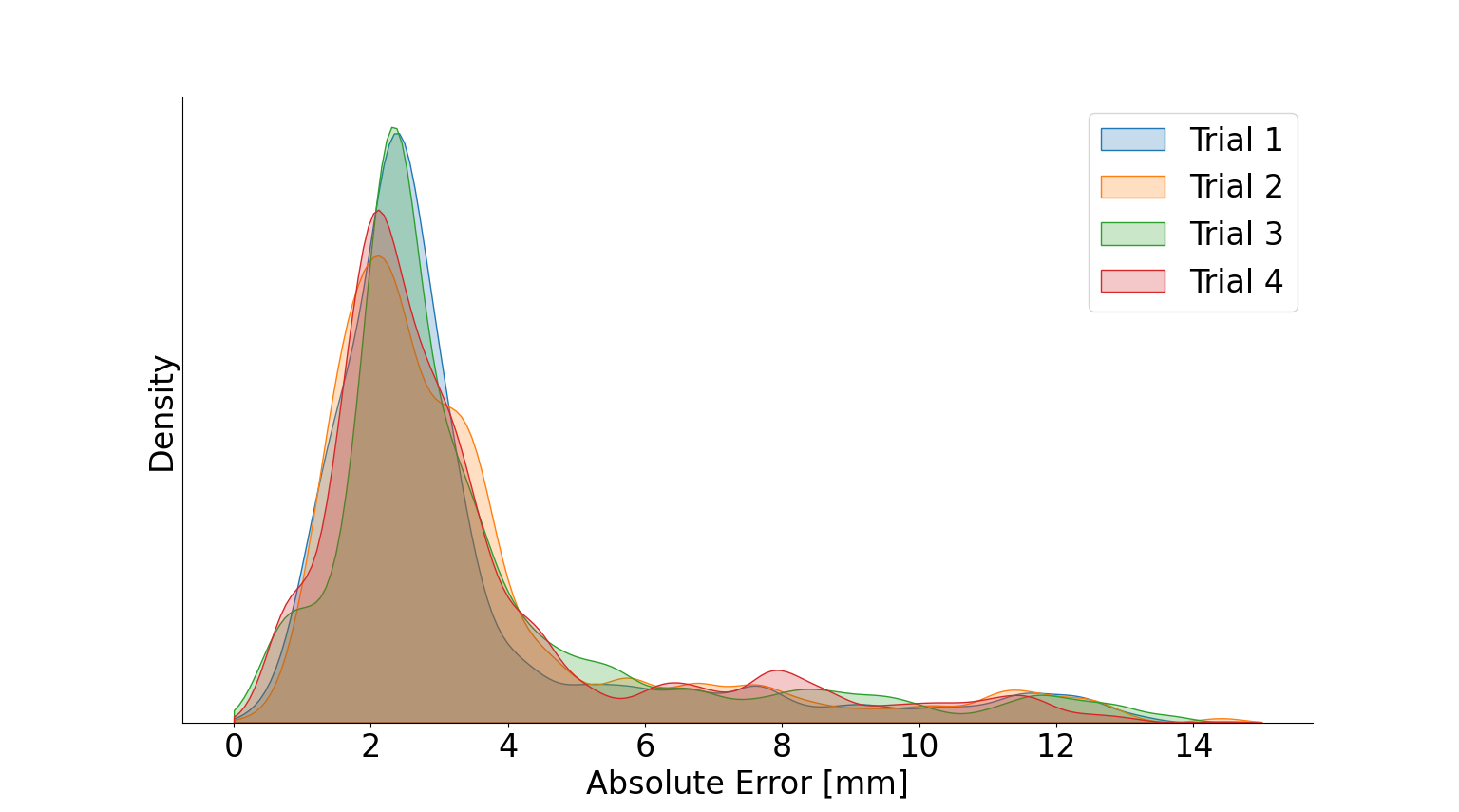}
        \caption{Density plots for absolute error across the four trials.}
        \label{fig::density_plot}
    \end{subfigure}
    \hfill
    \begin{subfigure}[b]{0.35\textwidth}
        \includegraphics[width=\textwidth]{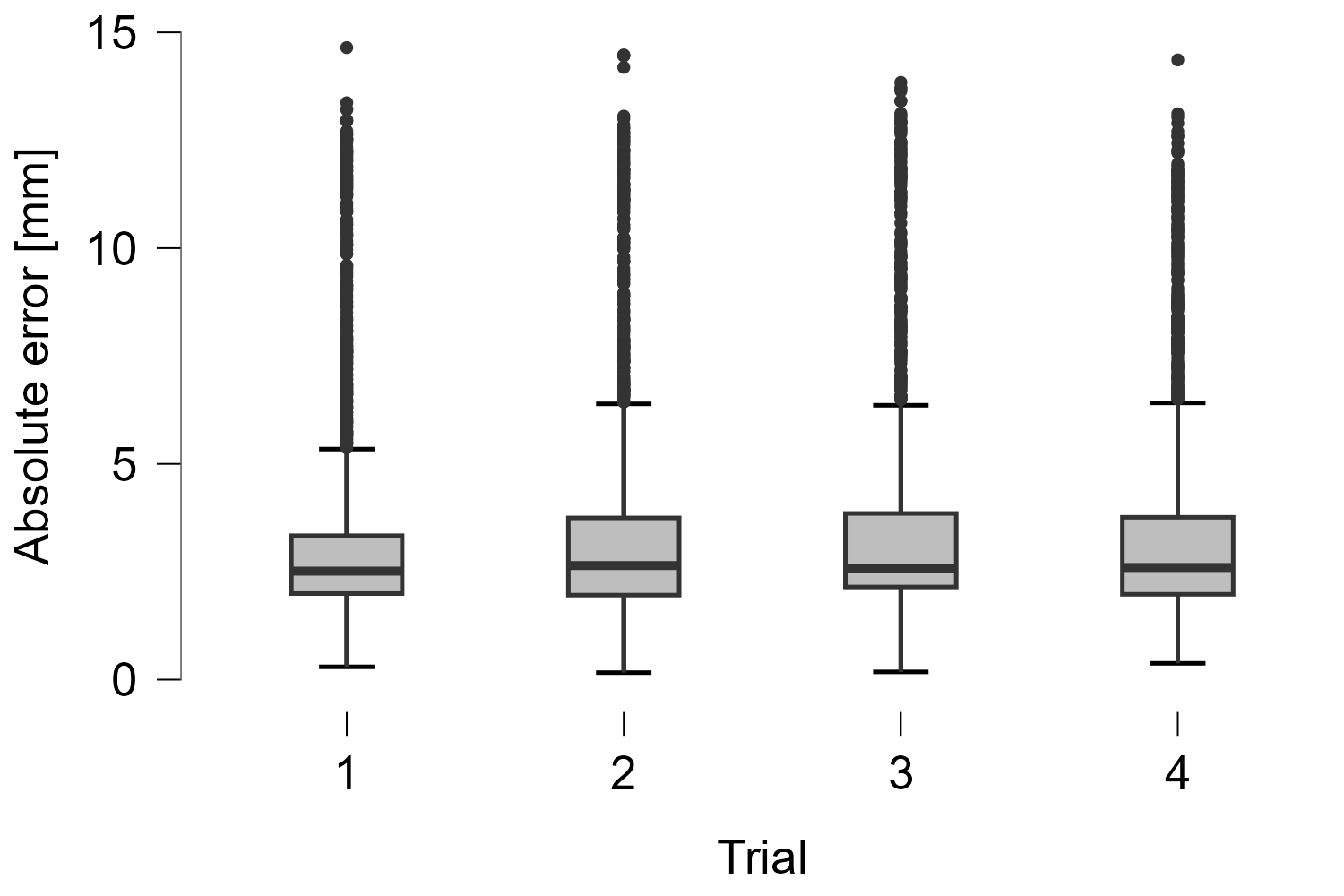}
        \caption{Box plots of 4 different trials.}
        \label{fig::box_plot}
    \end{subfigure}
    \caption{Comparison of density and box plots across four trials.}
    \label{fig::combined_plots}
\end{figure}

The distribution plots (Figure \ref{fig::density_plot}) for the absolute error across the four trials indicate a similar pattern, with most errors concentrated around 2-4 mm and a few outliers extending beyond 10 mm. The box plots (Figure \ref{fig::box_plot}) also show a consistent spread of data, with median values and interquartile ranges remaining stable across all trials.

Given that the Shapiro-Wilk test indicated non-normality ($p < 0.001$ for all trials), we employed the Kruskal-Wallis test, a non-parametric alternative to ANOVA, to compare the absolute error across the four trials. The Kruskal-Wallis test yielded a statistically significant result ($p < 0.001$), suggesting differences among the trials. However, Dunn's post-hoc test (Table \ref{tab:PostHocComparisonsTrial}) showed that most pairwise comparisons were not significant after adjusting for multiple comparisons (e.g., $p > 0.05$ for 1 vs. 2, 1 vs. 4, etc.), indicating that while statistical significance was detected overall, the practical differences between trials are minimal.

\begin{table}[h]
\centering
\begin{tabular}{lrrrrrr}
Comparison & z  & p & p${bonf}$ & p${holm}$ \\
\hline
1 - 2 & $-2.406$ & $0.016$ & $0.097$ & $0.064$ \\
1 - 3 & $-4.046$ & $<$ .001 & $<$ .001 & $<$ .001 \\
1 - 4 & $-1.335$  & $0.182$ & $1.000$ & $0.364$ \\
2 - 3 & $-1.640$  & $0.101$ & $0.606$ & $0.303$ \\
2 - 4 & $1.074$  & $0.283$ & $1.000$ & $0.364$ \\
3 - 4 & $2.715$ & $0.007$ & $0.040$ & $0.033$
\end{tabular}
\caption{Dunn's Post Hoc Comparisons - Trial}
\label{tab:PostHocComparisonsTrial}
\end{table}

Despite the statistical significance found in the Kruskal-Wallis test, the practical significance of the differences between trials is minimal, as supported by the visual analysis in Figure \ref{fig::combined_plots} and the Dunn’s post-hoc results (\ref{tab:PostHocComparisonsTrial}). This suggests that the observed variations in absolute error across repeated trials are consistent and do not significantly affect the reliability of the measurements. Therefore, the trials can be considered repeatable and consistent. This findings we apply also on all other cases.

\subsubsection{Varying Velocity and Displacement: Combination Analyses}

Given the observed impact of both velocity and displacement on tracker precision, we conducted a combined analysis using a two-way ANOVA to better understand their interactive effects. Despite the non-normality of the data, the large sample size of approximately 12,000 per case justifies the use of ANOVA in this context. As described in Section \ref{Procedure}, the experiment was constrained to a 3x3 grid of displacements ranging from 10 cm (Pattern B) to approximately 80x40x20 cm (Pattern D), with average velocities between 0.05 m/s and 0.2 m/s.

The \ref{tab:ANOVAAbsoluteError} presents the results of the two-way ANOVA, which examines the effects of velocity, displacement size, and their interaction on the absolute error.

\begin{table}[h]
	\centering
	{
		\begin{tabular}{lrrrrrrr}
			& df & Mean sq & p & $\eta$$^{2}$ & $\eta$$^{2}$  \\
			\hline
			velocity & $2$ & $576.254$ & $<$ .001 & $0.002$ & $0.005$  \\
			cube\_size  & $2$ & $103715.312$ & $<$ .001 & $0.415$ & $0.466$  \\
			velocity * cube\_size & $4$ & $584.015$ & $<$ .001 & $0.005$ & $0.010$  \\
		\end{tabular}
	}
	\caption{ANOVA - AbsoluteError [mm]}
	\label{tab:ANOVAAbsoluteError}
\end{table}

The results in Table \ref{tab:ANOVAAbsoluteError} indicate that both velocity and displacement size significantly affect the absolute error (p $<$ 0.001). Furthermore, the interaction between velocity and displacement size is also statistically significant (p $<$ 0.001). The effect sizes, represented by $\eta^2$ and $\eta^2_{p}$, show that displacement size has a much larger impact on absolute error (0.415 and 0.466, respectively) compared to velocity and their interaction. This observation aligns with our earlier findings, where the effect of velocity on tracking precision was primarily noticeable at velocities exceeding 30 cm/s.

Descriptive statistics for the absolute error across different velocities and displacement sizes are provided in Table \ref{tab:descriptives-AbsoluteError}.

\begin{table}[h]
	\centering
	{
		\begin{tabular}{lrrrrrrr}
    		\textbf{v (m/s)} & \textbf{Size (cm)} & \textbf{N} & \textbf{Mean} & \textbf{SD} & \textbf{SE} & \textbf{CoV} \\
			\hline
			0.05 & 10 & 12384 & 2.278 & 0.859 & 0.005 & 0.377 \\
			     & 20 & 12565 & 3.870 & 1.217 & 0.008 & 0.315 \\
			     & 40 & 12864 & 6.734 & 2.474 & 0.011 & 0.367 \\
			0.1  & 10 & 12156 & 2.922 & 1.094 & 0.006 & 0.374 \\
			     & 20 & 13570 & 3.975 & 1.512 & 0.007 & 0.380 \\
			     & 40 & 12403 & 6.757 & 2.271 & 0.014 & 0.336 \\
			0.2  & 10 & 11121 & 2.084 & 1.190 & 0.011 & 0.571 \\
			     & 20 & 12060 & 3.694 & 1.402 & 0.018 & 0.380 \\
			     & 40 & 13029 & 7.095 & 2.409 & 0.021 & 0.339 \\
		\end{tabular}
		\caption{Descriptive Statistics - Absolute Error [mm]}
	    \label{tab:descriptives-AbsoluteError}
	}
\end{table}

From Table \ref{tab:descriptives-AbsoluteError}, we observe the following:

1. \textbf{Effect of Velocity}: As the velocity increases from 0.05 m/s to 0.2 m/s, there is a general trend of increasing mean absolute error across all cube sizes. This suggests that higher velocities tend to decrease the tracking precision of the VIVE Ultimate Tracker, consistent with our previous observations.

2. \textbf{Effect of Displacement Size}: Larger displacement sizes consistently result in higher mean absolute errors. For example, at a velocity of 0.05 m/s, the mean absolute error increases from 2.278 mm at a 10 cm displacement to 6.734 mm at an 80x40x20 cm displacement.

3. \textbf{Interaction Effect}: The significant interaction effect between velocity and displacement size, as indicated by the ANOVA results, suggests that the combined increase in both velocity and displacement size leads to a more pronounced increase in absolute error. This is particularly evident at the highest velocity (0.2 m/s) and largest displacement (80x40x20 cm), where the mean absolute error reaches 7.095 mm.

4. \textbf{Coefficient of Variation}: The coefficient of variation remains relatively stable across different conditions, indicating consistent variability in the measurements relative to the mean error.

In summary, both velocity and displacement significantly impact the precision of the VIVE Ultimate Tracker. However, due to the robot's speed limitations, we were unable to fully explore the velocity effects beyond 30 cm/s, where these effects appear to be more critical. This would explain the rather weak combination effect of displacement and velocity.

\end{document}